\documentclass{article}

\usepackage{microtype}
\usepackage{graphicx}
\usepackage{subcaption}
\usepackage{booktabs} % for professional tables
\usepackage{amsmath}
\usepackage{amssymb}
\usepackage{bm}
\usepackage{multirow}
\usepackage{caption}
\usepackage{tabularx}
\usepackage{hyperref}

\usepackage[preprint]{icml2026}

\usepackage{amsmath}
\usepackage{amssymb}
\usepackage{mathtools}
\usepackage{amsthm}
\usepackage{tcolorbox}
\usepackage{enumitem}

\usepackage[capitalize,noabbrev]{cleveref}

\theoremstyle{plain}

\theoremstyle{definition}

\theoremstyle{remark}

\usepackage[textsize=tiny]{todonotes}

\icmltitlerunning{VIRAL: Visual In-Context Reasoning via Analogy in Diffusion Transformers}

\begin{document}

\twocolumn[
  \icmltitle{VIRAL: Visual In-Context Reasoning via Analogy in Diffusion Transformers}



  \icmlsetsymbol{equal}{*}

  \begin{icmlauthorlist}
    \icmlauthor{Zhiwen Li}{ecnu}
    \icmlauthor{Zhongjie Duan}{alibaba}
    \icmlauthor{Jinyan Ye}{ecnu}
    \icmlauthor{Cen Chen}{ecnu}
    \icmlauthor{Daoyuan Chen}{alibaba}
    \icmlauthor{Yaliang Li}{alibaba}
    \icmlauthor{Yingda Chen}{alibaba}
  \end{icmlauthorlist}

  \icmlaffiliation{ecnu}{School of Data Science and Engineering, East China Normal University, Shanghai, China}
  \icmlaffiliation{alibaba}{Alibaba Group, Hangzhou, China}

  \icmlcorrespondingauthor{Cen Chen}{cenchen@dase.ecnu.edu.cn}

  \icmlkeywords{Machine Learning, ICML}

  \begin{center}
    \vskip 0.1in
    \centerline{\includegraphics[width=0.94\textwidth]{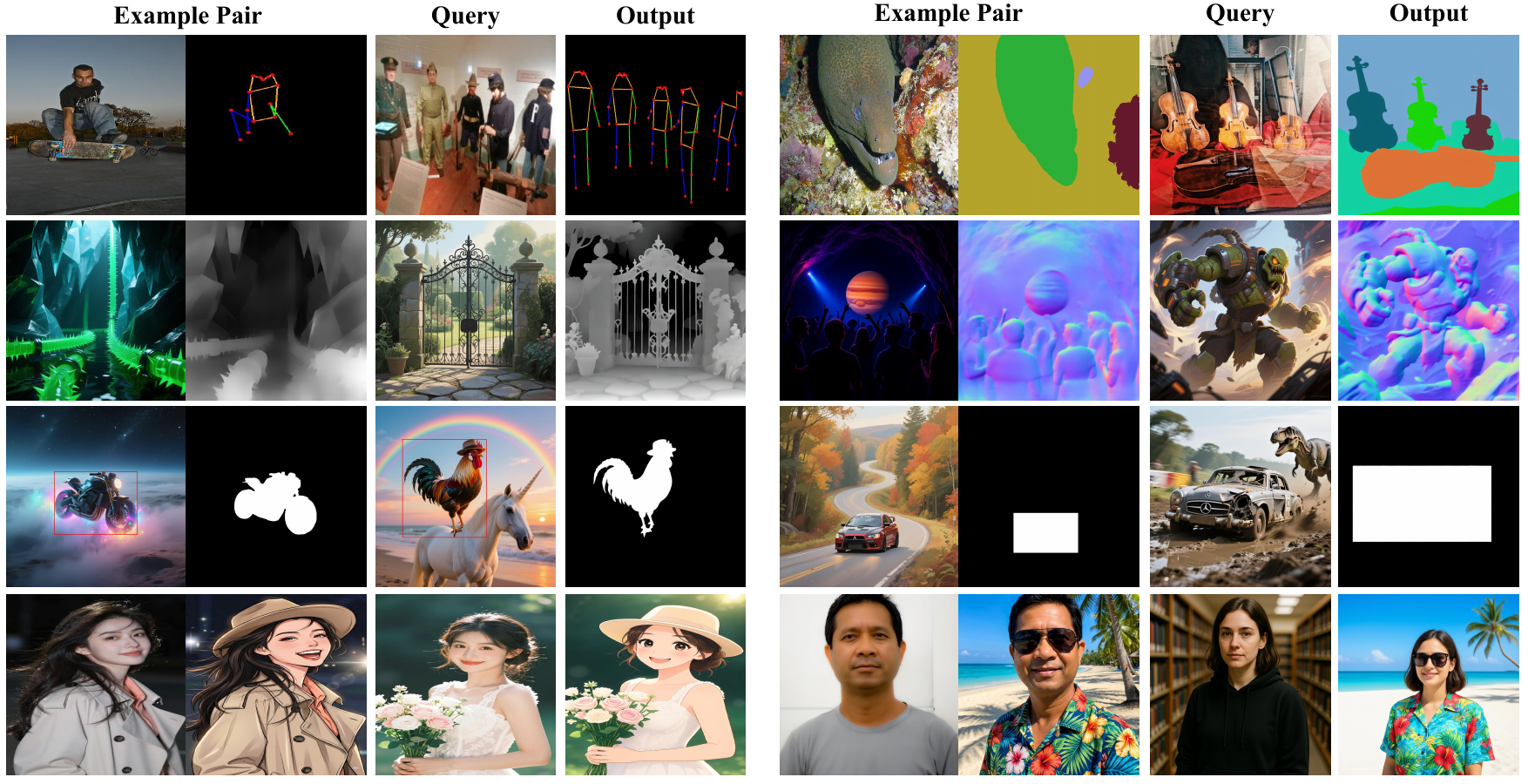}} % 替换你的文件名
    % 调整图片与Caption的距离
    \captionof{figure}{Illustration of in-context learning with \textbf{VIRAL}. Given a reference exemplar pair, VIRAL interprets the underlying visual transformation and applies it to a query image, including  standard visual task and open-domain editing.}
    \label{fig:teaser}
  \end{center}
  \vskip 0.3in
]

\printAffiliationsAndNotice{}  % no special notice (required even if empty)

\begin{abstract}
Replicating In-Context Learning (ICL) in computer vision remains challenging due to task heterogeneity. We propose \textbf{VIRAL}, a framework that elicits visual reasoning from a pre-trained image editing model by formulating ICL as conditional generation via visual analogy ($x_s : x_t :: x_q : y_q$). We adapt a frozen Diffusion Transformer (DiT) using role-aware multi-image conditioning and introduce a Mixture-of-Experts LoRA to mitigate gradient interference across diverse tasks. Additionally, to bridge the gaps in current visual context datasets, we curate a large-scale dataset spanning perception, restoration, and editing. Experiments demonstrate that VIRAL outperforms existing methods, validating that a unified V-ICL paradigm can handle the majority of visual tasks, including open-domain editing. Our code is available at \url{https://anonymous.4open.science/r/VIRAL-744A}
\end{abstract}

\section{Introduction}

In-context learning (ICL) has become a powerful paradigm in the field of natural language processing. It enables pre-trained models to infer potential input-output mappings from a small number of examples and apply them to new queries without requiring task-specific parameters or fine-tuning \cite{Tom2020gpt3,yaru2022llmGeneralPurpose,hugo2023llama}. Following the success of ICL in NLP, visual in-context learning (V-ICL) has also gradually attracted attention from academia in recent years \cite{Amir2022visualprompt,Weihuang2023Explicit,Xinlong2023SegGPT} .
% For example, recent studies \cite{Amir2022visualprompt,Weihuang2023Explicit,Xinlong2023SegGPT} have made promising attempts to utilize in-context capabilities to complete some downstream computer vision tasks.

Despite the promising future of V-ICL, it still faces challenges in practice. A fundamental challenge stems from the heterogeneity of visual transformations, different visual tasks have vastly different input and output representations, often requiring task-specific loss functions and architectural designs. This necessitates that implementing V-ICL on a single model first requires the model to possess various basic task knowledge, such as depth map \cite{Lihe2024depthany} and normal map \cite{Chongjie2024stalenormal}. Further, achieving broader open-domain editing requires even stronger semantic prior knowledge. 
Another key bottleneck in V-ICL development is the current lack of high-quality image in-context datasets, as the model requires exposure to diverse exemplar-query quadruplets to effectively decouple the underlying transformation from specific visual content. Moreover, most existing V-ICL approaches \cite{Jiarui2024IMProv,Xinlong2023painter,Amir2022visualprompt} typically rely on training image inpainting models from scratch on grid-structured image datasets. During inference, they concatenate example image pairs and query images into a grid-like canvas, using a placeholder mask to represent the target image to be predicted. The model then inpaints the prediction based on this grid input. This stitching method limits the resolution and semantic expressive power of individual images, often proving effective only on a few specific tasks and ineffective for open-domain editing. And this paradigm do not to leverage the power of modern pre-trained visual foundational models (such as SD \cite{Robin2022ldm} and Qwen-Image \cite{wu2025qwenimagetechnicalreport}) and overlooks a major advantage of ICL.

We start in the observation that a broad spectrum of vision tasks fundamentally operates as an image-to-image transformation. For instance, semantic segmentation effectively recasts a natural image into a mask visualization, whereas image restoration tasks like deraining recover a clean signal from a degraded input. This shared structure motivates a unified V-ICL interface where the model should infer the specific transformation logic from an exemplar pair $(x_s \rightarrow x_t)$ and apply it to a new query image $x_q$.

Guided by this view, we propose to elicit visual in-context reasoning capabilities directly from a pre-trained DiT-based image editing model rather than training a new generalist architecture from scratch. Specifically, we formulate V-ICL as a visual analogy conditional generation task ($x_s : x_t :: x_q : y_q$) grounded in a unified RGB image space. To facilitate this inference, we adapt a DiT-based editing backbone to process multi-image visual context via a role-aware token sequence. Furthermore, we implement parameter-efficient multi-task adaptation using a Mixture-of-Experts LoRA (MoE-LoRA), a strategy that effectively mitigates interference across heterogeneous tasks while preserving the generative priors of the frozen backbone.

To bridge the gap in existing datasets, we construct a comprehensive in-context image editing dataset that spans a broad spectrum of visual reasoning tasks, ranging from standard perception and restoration tasks, such as segmentation visualization and low-light enhancement, to open-domain editing scenarios facilitated by analogous quadruplets mined and synthesized from instruction-driven corpora.

Empirically, our model demonstrates robust cross-task generalization by performing a diverse array of downstream tasks given a single visual demonstration, as shown in Figure \ref{fig:teaser}. 
% Furthermore, VIRAL outperforms prior V-ICL baselines across both structured perception and photorealistic restoration domains, achieving performance that rivals or even surpasses specialized models on many tasks. These results serve as compelling evidence that effective in-context adaptation can be elicited directly from a pre-trained DiT backbone.
Furthermore, VIRAL outperforms previous V-ICL baseline models on a variety of vision tasks, achieving performance comparable to or even surpassing professional models. These results strongly demonstrate that effective context adaptation capabilities can be directly derived from pre-trained DiT backbone networks.

In summary, our main contributions are as follow:
\begin{itemize}
    \item \textbf{A unified generative formulation of V-ICL.} We introduce \textbf{VIRAL}, a framework that recasts V-ICL as a visual analogy conditional generation task within a continuous RGB space, establishing a universal generative interface that seamlessly adapts to perception, restoration, and open-ended editing.
    \item \textbf{Empirical validation of universal V-ICL feasibility.} We demonstrate that unifying diverse tasks into a generative format enables pre-trained models to effectively perform visual in-context learning.
    \item \textbf{A large-scale in-context editing dataset.} We construct and open-source a comprehensive dataset of exemplar-query quadruplets, covering a spectrum from standard visual transformations to open-domain edits.
\end{itemize}

\section{Related Work}

\paragraph{In-Context Learning.}
The emergence of Large Language Models (LLMs), especially GPT-3 \cite{Tom2020gpt3}, has revolutionized the paradigm of natural language processing by introducing In-Context Learning (ICL). Unlike traditional fine-tuning, which requires updating gradients for each downstream task, ICL enables models to infer task objectives from a small number of examples provided in the prompts, allowing them to adapt to new tasks effectively and promptly without gradient updates or fine-tuning \cite{Aakanksha2023palm,Jason2022Emergent}. Recent research \cite{Yingcong2023TransformersasAlgorithms} points out that ICL is essentially an optimization algorithm, and the Transformer is actually learning how to "optimize." Through large-scale pre-training, it learns a set of general optimization strategies, enabling it to quickly adapt to new tasks during the inference phase. 
Shared Transformer architectures imply that modern generative models inherently possess contextual reasoning potential.

\paragraph{Visual In-Context Learning.}
The computer vision community has sought to replicate the In-Context Learning effect within the visual domain.  Early pioneers, such as VisualPrompt \cite{Amir2022visualprompt} and IMProv \cite{Jiarui2024IMProv}, trained ViT-based MAE-VQGAN models \cite{Kaiming2022Masked} to reformulate heterogeneous vision tasks as image inpainting problems. These models were trained on uncurated datasets consisting of structured, document-style imagery and employed grid-like structures with placeholder masks to facilitate prediction during inference. Painter \cite{Xinlong2023painter} extends this paradigm by leveraging large-scale, annotated task-specific image pairs instead of uncurated data. While these methods demonstrated initial feasibility, they frequently encountered bottlenecks, including imprecise contextual inference, limited spatial resolution, and poor generalization to high-level semantic tasks.

With the emergence of Latent Diffusion Models (LDMs) \cite{Robin2022ldm}, works such as Prompt Diffusion \cite{Zhendong2023promptdiffusion} attempted to inject contextual conditions via ad-hoc ControlNet \cite{Lvmin2023controlnet} branches and specialized image encoders. However, such methods remain restricted to a narrow set of predefined tasks. Conversely, SD-VICL \cite{Trevine2025sdvicl} introduced a training-free ICL mechanism by manipulating the internal cross-attention maps of Stable Diffusion—specifically by utilizing reference pairs as Key and Value states for the Query image. Nevertheless, this approach necessitates computationally expensive image inversion and imposes strict semantic alignment constraints between the query and the reference. And the carefully designed attention mechanism of the UNet structure for SD \cite{Robin2022ldm} cannot be adapted to other model architectures.
\begin{figure*}[t]
    \centering
    \includegraphics[width=0.96\linewidth]{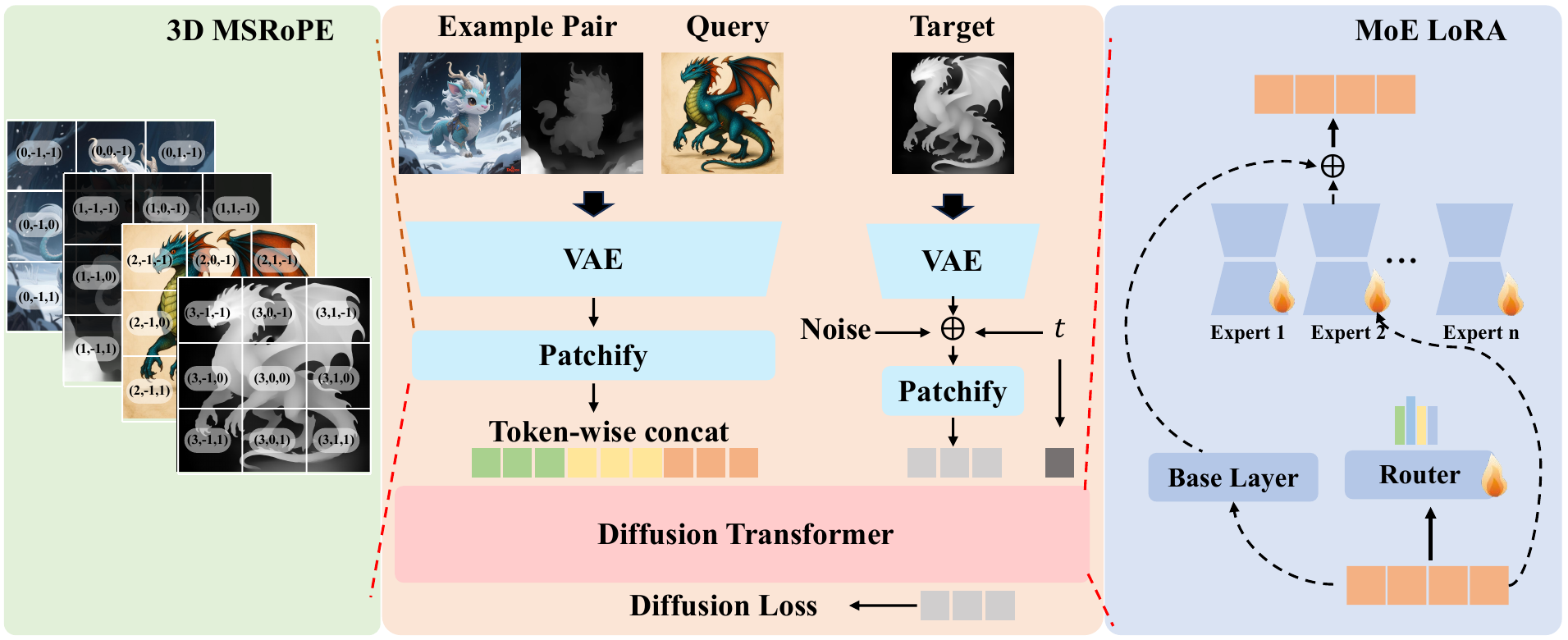}
    \caption{\textbf{Overview of the proposed Visual In-Context Learning framework.} 
We unify diverse visual tasks into a homogeneous RGB pixel space, enabling a universal generative interface. 
The visual tokens from the reference exemplar pair  and the query image are concatenated along the sequence dimension and fed into the Diffusion Transformer (DiT) backbone. The exemplar pair and the query images remain fixed during denoising steps, while the model updates only the noisy latent to the target image.}
    \label{fig:framework}
\end{figure*}
Unlike previous studies, we adopt a LoRA-adapted model on a unified DiT backbone architecture and effectively address the heterogeneity problem of visual tasks, enabling it to serve as a potential general learner that seamlessly integrates low-level image inpainting, high-level perception, and creative editi.

\section{Method}
\label{sec:method}

\subsection{Problem Setup: Visual Analogy Conditional Generation}
\label{sec:problem}
We formally define the single-shot V-ICL setting.
Let $\mathcal{X} \subseteq \mathbb{R}^{H \times W \times 3}$ denote the RGB image space.
Each visual task is governed by an underlying transformation operator $\mathcal{T} \colon \mathcal{X} \rightarrow \mathcal{X}$.
This transformation is exemplified by a support pair $(x_s, x_t)$, such that:
\begin{equation}
    x_t = \mathcal{T}(x_s).
\end{equation}
Given a query source image $x_q$, our objective is to synthesize a target $\hat{y}_q$ that approximates the ground truth transformation:
\begin{equation}
    \hat{y}_q \approx \mathcal{T}(x_q).
\end{equation}
Crucially, this generation must be performed without task-specific heads and without test-time parameter updates.
We formulate this objective as solving a visual analogy problem:
\begin{equation}
    x_s : x_t  ::  x_q : \hat{y}_q.
\end{equation}
While text instructions $I$ are optionally available in some settings, we adhere to the strict regime where $I=\emptyset$.
Consequently, the model must infer the intended transformation $\mathcal{T}$ solely from the visual correlation within the exemplar pair.
An overview of our framework is shown in Figure \ref{fig:framework}.

\subsection{Backbone: Pre-trained Image Editing Model}
Our framework leverages a pre-trained image editing architecture comprising a Diffusion Transformer (DiT) as the denoising backbone. This foundation encapsulates robust generative priors and extensive general visual knowledge, which are essential for high-fidelity synthesis.
% Adhering to a minimalist design philosophy, we maintain the structural integrity of the backbone.
Instead of architectural modifications, we enable in-context inference by seamlessly injecting exemplar pairs as conditioning tokens and employing parameter-efficient adaptation strategies to align the model with the visual analogy objective.

\subsection{Role-aware Multi-image Token Conditioning}
\label{sec:conditioning}

\paragraph{Latent tokens.}
Let $\mathrm{Enc}(\cdot)$ denote the composite operation of the frozen VAE encoder~\cite{Diederik2014vae} followed by patchification. This function maps an input RGB image to a sequence of $L$ visual tokens residing in $\mathbb{R}^{L \times D}$. Accordingly, we encode the exemplar source, exemplar target, and query source images into their respective latent token representations:
\begin{equation}
    \mathbf{z}_s = \mathrm{Enc}(x_s), \quad
    \mathbf{z}_t = \mathrm{Enc}(x_t), \quad
    \mathbf{z}_q = \mathrm{Enc}(x_q).
\end{equation}

\paragraph{Condition sequence.}
To construct the holistic visual context, we concatenate the latent tokens of the exemplar pair and the query source along the sequence dimension, yielding a unified conditioning tensor:
\begin{equation}
    \mathbf{Z}_{\mathrm{cond}} = \mathrm{Concat}(\mathbf{z}_s, \mathbf{z}_t, \mathbf{z}_q) \in \mathbb{R}^{3L \times D}.
\end{equation}

\paragraph{Role and position encoding.}
To distinguish token roles across images (Figure~\ref{fig:framework}), we employ a 3D-MSRoPE strategy. Extending the MSRoPE mechanism from Qwen-Image~\cite{wu2025qwenimagetechnicalreport}, we incorporate an additional topological dimension to explicitly encode ICL roles. This design preserves intra-image spatial geometry while establishing distinct inter-image identities, thereby enabling precise global cross-attention for transformation inference.

\subsection{Diffusion Training Objective with In-context Conditioning}
\label{sec:diffusion}

Let $y_q$ denote the ground-truth target for the query image. Following standard diffusion dynamics, we perturb the latent representation of $y_q$ to an arbitrary timestep $t$, yielding the noisy state $\mathbf{z}_{y,t}$. The DiT backbone then estimates the noise component (or flow velocity) $\hat{\epsilon}_\theta$, conditioned on the noisy latent and the unified context sequence $\mathbf{Z}_{\mathrm{cond}}$:
\begin{equation}
    \hat{\epsilon}_\theta = \mathrm{DiT}_\theta(\mathbf{z}_{y,t}, t \mid \mathbf{Z}_{\mathrm{cond}}).
\end{equation}
We optimize the standard objective function over the data distribution:
\begin{equation}
    \mathcal{L}(\theta) = \mathbb{E}_{(x_s,x_t,x_q,y_q),t}\left[\; \ell(\hat{\epsilon}_\theta, \epsilon) \;\right],
\end{equation}
where $\ell$ denotes the loss function (e.g., MSE) and $\theta$ represents the trainable parameters (including adapters). During denoising stage, the conditioning context $\mathbf{Z}_{\mathrm{cond}}$ remains stationary, guiding the iterative denoising process from Gaussian noise to the final edited output $\hat{y}_q$.

We train on a dataset of exemplar-query quadruplets
\begin{equation}
    \mathcal{Q} = \big\{(x_s, x_t), (x_q, y_q)\big\},
\end{equation}
where both pairs share the same underlying transformation $\mathcal{T}$ but differ in content and scenes. Section~\ref{sec:data_construction} details the sources, filtering, and quality control.

\subsection{MoE-LoRA for Heterogeneous In-context Tasks}
\label{sec:moelora}

To mitigate the potential gradient interference arising from diverse visual tasks, we enhance the standard LoRA with a Mixture-of-Experts (MoE) formulation \cite{Albert2024moe}, selectively applying it to the DiT layers.
Formally, given a frozen projection $W_{\mathrm{base}}$, we introduce $N$ LoRA experts $\{E_i\}_{i=1}^N$. The layer output $h$ is the weighted sum of the base projection and the top-$k$ active experts:
\begin{equation}
    h = W_{\mathrm{base}}x + \sum_{i \in \mathcal{S}} g_i(x) \cdot (B_i A_i x),
\end{equation}
where $A_i, B_i$ are low-rank matrices. The gating weights $g(x)$ and the active expert set $\mathcal{S}$ are determined by a differentiable router $W_g$.  Specifically, we select the Top-$k$ active experts via the following routing mechanism::
\begin{equation}
    g(x) = \operatorname{Softmax}(W_g x), \quad \mathcal{S} = \operatorname{TopK}(g(x), k).
\end{equation}
To prevent mode collapse and ensure uniform expert utilization, we introduce an auxiliary load-balancing loss $\mathcal{L}_{aux}$:
\begin{equation}
    \mathcal{L}_{aux} = N \sum_{i=1}^N f_i \cdot \bar{P}_i,
\end{equation}
where $f_i$ is the fraction of tokens assigned to expert $i$ in a batch, and $\bar{P}_i$ is the average routing probability for expert $i$. 

\section{Visual In-Context Dataset}
\label{sec:data_construction}

% To operationalize the unified training formulation outlined in Section~\ref{sec:diffusion}, 
We construct a comprehensive dataset of exemplar-query quadruplets. Our construction pipeline is organized into two streams based on the nature of the transformation $\mathcal{T}$: \textbf{Standard Visual Tasks}, where the transformation logic is predefined and globally consistent; and \textbf{Open-domain Editing}, where $\mathcal{T}$ is unstructured. More details and samples of dataset are provided in Appendix \ref{sec:appendix_dataset_params} and \ref{sec:dataset_sample}

\paragraph{Standard Visual Tasks.}
To establish a robust foundational corpus, we construct a large-scale self-generated dataset. Specifically, we sample diverse text prompts from DiffusionDB~\cite{wangDiffusionDBLargescalePrompt2022} and utilize Qwen-Image to synthesize high-fidelity source images. We then employ an automated annotation pipeline to generate paired ground truths: ControlNetAux \cite{controlnet_aux} is applied to produce dense edge, depth, and surface normal maps, while Qwen2.5-VL-7B-Instruct is leveraged to identify salient entities, providing precise category labels and bounding box annotations. Here, any two randomly selected instances for a specific task naturally constitute a valid training quadruple, the specific input ($x$) and output ($y$) formulations are defined as follows:

\begin{itemize}
    \item \textbf{Dense Prediction and Spatial Localization:} 
    We primarily leverage the self-generate dataset. 
    For edge detection, depth estimation, and surface normal estimation, we directly utilize the paired RGB images and their corresponding ground-truth maps provided by the dataset. 
    For object detection, we select a subset of common categories and reformulate the task as category-specific localization. The target $y$ is rendered as a binary mask featuring a filled white rectangle on a black canvas based on the bounding box annotations. Crucially, we enforce category consistency within each quadruplet, ensuring that both the exemplar and the query target the same object class.
    Similarly, for people keypoints detection, we employ samples from COCO-2017~\cite{Tsung2014coco}, rendering the skeletal annotations into visual pose maps as the target $y$.
    
    \item \textbf{Segmentation Tasks:} We address both interactive and entity-level scenarios. 
    For interactive segmentation, we simulate user-specified selection by superimposing a visible red bounding box around a target object on the source image $x$. The target $y$ is a  binary mask, generated by prompting the Segment Anything Model (SAM)~\cite{kirillov2023segany} with the object's ground-truth box coordinates.
    For \textit{entity segmentation}, we adopt the EntityV2~\cite{qi2022fine} dataset 
    % To distinguish individual instances without semantic constraints, 
    and map categorical masks to random distinct colors, constructing a panoptic-style visualization as the target $y$.
    
    \item \textbf{Image Restoration and Enhancement:} We formulate these tasks as inverse problems, mapping a degraded input $x$ to a high-fidelity reference $y$. 
    For \textit{colorization}, we synthesize the input $x$ by desaturating the RGB target $y$. 
    For \textit{watermark removal}, we generate training pairs by superimposing logo templates from CLWD~\cite{Yang2021WDNet} onto clean images to create the watermarked source $x$. 
    Additionally, for \textit{deraining} and \textit{low-light enhancement}, we adopt Rain200L~\cite{Wenhan2017rain200l} and LoLv2~\cite{Wenhan2021lolv2}, where the provided rainy or low-light images serve as the input $x$ and their clean counterparts as the target $y$.
\end{itemize}

\paragraph{Open-domain In-Context Editing Dataset.}
% To empower the model with generic in-context reasoning capabilities beyond predefined tasks, we curate a large-scale dataset consisting of analogous task quadruplets.
% Unlike standard tasks, open-domain editing lacks a fixed label space. We must actively construct analogous quadruplets where the specific ``edit'' is consistent. 
To extend in-context reasoning beyond fixed taxonomies, we curate a large-scale dataset of analogous editing quadruplets. Given the unbounded semantic space of open-domain editing, ensuring transformation consistency across pairs is critical. To address this, we implement two complementary strategies that exploit existing instruction-driven corpora, such as GPT-Image-Edit-1.5M~\cite{Yuhan2025gptimgedit} and Pico-Banana-400k~\cite{Yusu2025pico}, to source valid quadruplets.

\textbf{Generative Analogous Synthesis.}
We implement a pipeline leveraging an LLM and a Text-to-Image model to synthesize analogous editing pairs. Starting with a valid reference tuple $(x_1, y_1)$ and its associated editing instruction $I$, we prompt the LLM to generate a description $c_{new}$ for a semantically distinct scene that remains compatible with $I$. This caption is fed into Qwen-Image~\cite{wu2025qwenimagetechnicalreport} to synthesize a novel source image $x_2$. Subsequently, we apply the original instruction $I$ to $x_2$ via Qwen-Image-Edit to produce the corresponding target $y_2$. This procedure yields a synthetic quadruplet, ensuring task consistency by applying the identical instruction $I$ to both scenes.

\textbf{Embedding-Space Task Mining.}
To uncover implicit analogous relationships within existing datasets, we devise a clustering-based retrieval framework. We hypothesize that the semantic transformation of an editing task can be modeled as a linear translation vector in the latent space. For a sample pair, the task vector $\mathbf{v}_{task}$ is defined as:
\begin{equation}
    \mathbf{v}_{task} = \mathcal{E}_{\text{CLIP}}(y) - \mathcal{E}_{\text{CLIP}}(x)
\end{equation}
where $\mathcal{E}_{\text{CLIP}}(\cdot)$ denotes the pre-trained CLIP ViT-L/14~\cite{Alec2021clip}. 
We apply K-Means clustering to these task vectors to aggregate samples exhibiting similar editing logic. Within each cluster, for a reference pair $P_i$, we retrieve its nearest neighbor $P_j$ based on cosine similarity to construct a candidate quadruplet. To ensure data quality, we enforce a dual-filtering strategy that eliminates visually redundant source images to prevent trivial mappings, while simultaneously verifying high textual similarity between instructions to guarantee semantic consistency.
\begin{figure*}[t!]
    \centering
\includegraphics[width=0.96\linewidth]{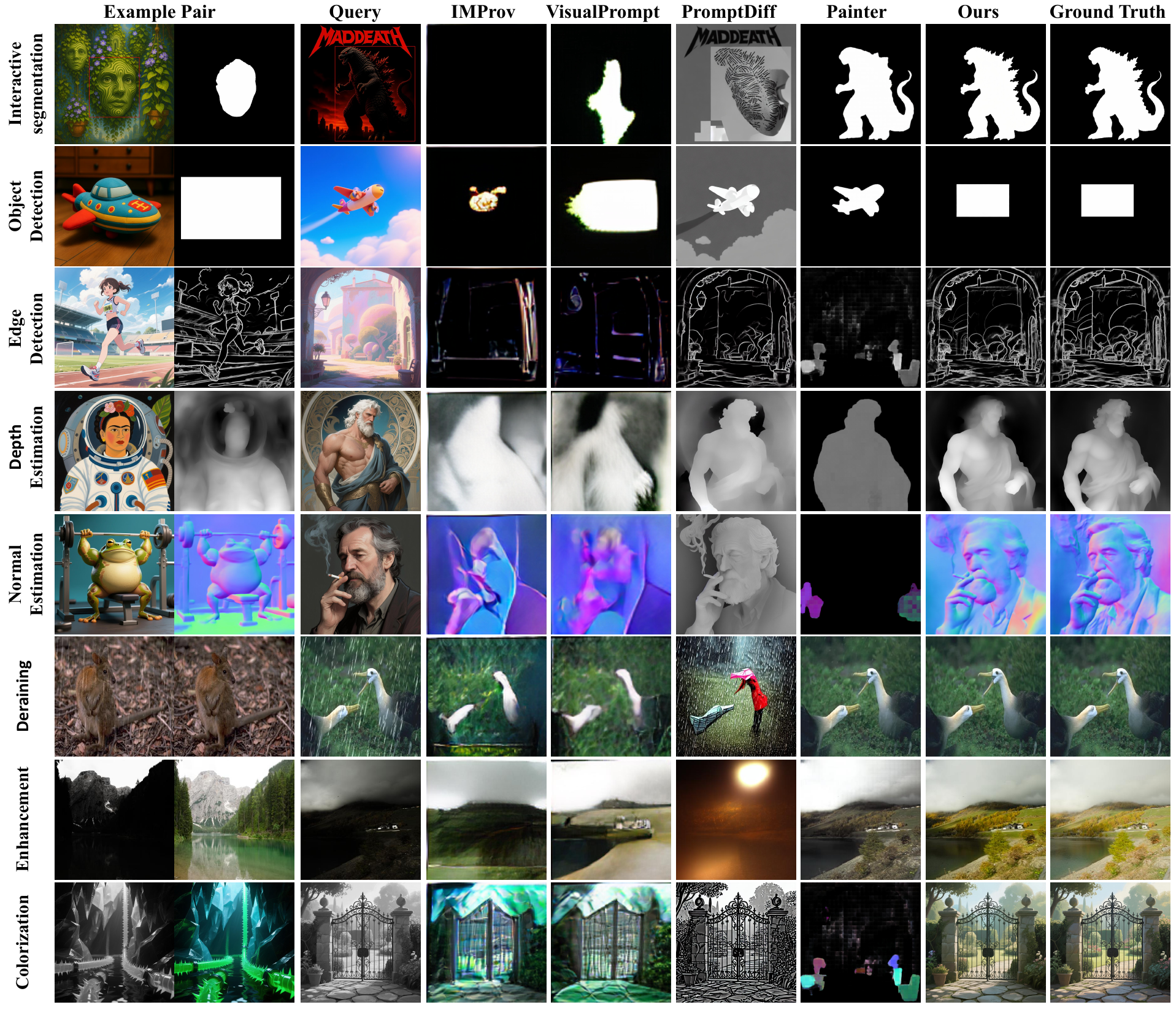}
    \caption{Quantitative comparison.
We evaluate the performance of four V-ICL baselines against VIRAL. 
While existing baselines either exhibit restricted task versatility or suffer from performance degradation when encountering complex scenarios due to their reliance on over-simplified training distributions, our model consistently achieves superior accuracy and visual fidelity across all evaluated tasks.}
    \label{fig:tasks_visual}
\end{figure*}
\paragraph{Data Statistics and Distribution.}
% The final dataset is rigorously curated to ensure comprehensive multi-task coverage. 
We leverage a shared pool of 100K image pairs to support depth estimation, surface normal prediction, edge detection, colorization, and watermark removal. This foundation is supplemented by 100K interactive segmentation pairs, 50K human pose samples, 20K entity segmentation pairs, and 8K object detection samples. Additionally, we incorporate 2K pairs each for deraining and low-light enhancement, alongside 40K open-domain editing quadruplets. For each task, a small, independent subset is strictly reserved for evaluation and remains unseen during the training phase.

\section{Experiment}

\subsection{Implementation Details}
We implement VIRAL based on the pre-trained Qwen-Image-Edit-2511~\cite{wu2025qwenimagetechnicalreport}. We inject MoE-LoRA modules ($N=4$, Top-2 routing) into the FFN layers of the DiT backbone, while standard LoRA is applied to other layers. The model is fine-tuned on our In-Context Dataset (Sec.~\ref{sec:data_construction}) and all quantitative evaluations are conducted on a held-out test set unseen during training. During inference, we adopt a 1-shot setting by providing a single task-specific exemplar pair. More details ablation studies on model design are provided in Appendix~\ref{sec:appendix_impl} and \ref{sec:c2}.

\subsection{Comparison with V-ICL Models}

\begin{table*}[t]
\centering
\small
\caption{Quantitative comparison results with the V-ICL model. ``Seg.'' refers to interactive segmentation.}
\label{tab:comparison}
\setlength{\tabcolsep}{1.9pt}
{\fontsize{8.5pt}{11pt}\selectfont
\begin{tabular}{c|c|c|cc|cc|cc|cc|cc|cc} 
\toprule
\multirow{2}{*}{\textbf{Method}} & \textbf{Seg.}& \textbf{Obj. Det.} & \multicolumn{2}{c|}{\textbf{Edge}} & \multicolumn{2}{c|}{\textbf{Colorization}} & \multicolumn{2}{c|}{\textbf{Depth}}             & \multicolumn{2}{c|}{\textbf{Normal}}        & \multicolumn{2}{c|}{\textbf{Deraining}} & \multicolumn{2}{c}{\textbf{Enhancement}}  \\
                        & IoU $\uparrow$               & IoU $\uparrow$           & RMSE $\downarrow$ & LPIPS $\downarrow$            & FID $\downarrow$   & LPIPS $\downarrow$                   & AbsRel $\downarrow$         & $\delta_1 \uparrow$ & Med $\downarrow$            & Mean $\downarrow$            & PSNR $\uparrow$ & SSIM $\uparrow$               & PSNR $\uparrow$           & SSIM $\uparrow$      \\ 
\midrule
IMProv                   & 0.178             & 0.337          & 99.36 & 0.682             & 210.89 & 0.714                    & 0.175          & 0.711                 & 56.08           & 52.77            & 15.29 & 0.330               & 15.14           & 0.353      \\
VisualPrompt            & 0.191             & 0.347          & 63.23 & 0.585             & 214.90 & 0.707                    & 0.173          & 0.732                 & 48.52           & 46.03            & 15.08 & 0.329               & 15.30           & 0.444      \\
PromptDiff             & 0.178             & 0.324          & 35.88 & 0.255             & 179.21 & 0.598                    & 0.160          & 0.714                 & 97.27           & 105.92           & 8.67  & 0.128               & 8.75            & 0.323      \\
Painter                 & 0.348             & 0.387          & 94.93 & 0.632             & 176.71 & 0.624                    & 0.167          & 0.732                 & 116.01          & 119.87           & 19.13 & 0.634               & 16.39          & 0.712      \\
\textbf{Ours}           & \textbf{0.795  }           & \textbf{0.562}          & \textbf{28.59} & \textbf{0.133}             & \textbf{43.75}  & \textbf{0.126}                    & \textbf{0.138} & \textbf{0.812}        & \textbf{17.582} & \textbf{ 20.187} & \textbf{29.67} & \textbf{0.889 }              & \textbf{25.24} & \textbf{0.878}      \\
\bottomrule
\end{tabular}
}
\end{table*}

We quantitatively evaluate VIRAL against representative Visual In-Context Learning (V-ICL) baselines, including Painter~\cite{Xinlong2023painter}, IMProv~\cite{Jiarui2024IMProv}, VisualPrompt~\cite{Amir2022visualprompt}, and PromptDiff~\cite{Zhendong2023promptdiffusion}. To ensure a comprehensive assessment, the benchmark spans diverse visual reasoning tasks ranging from high-level perception to low-level restoration. Specifically, we follow the evaluation protocols of PromptDiff~\cite{Zhendong2023promptdiffusion} for edge detection, SD-VICL~\cite{Trevine2025sdvicl} for colorization, and adopt standard metrics from DepthAnything~\cite{Lihe2024depthany} and StableNormal~\cite{Chongjie2024stalenormal} for geometric estimation (depth and normal). For image restoration, we employ the test pipelines from CSUD~\cite{Guanglu2025csud} and HVID~\cite{Qingsen2025hvi} for deraining and low-light enhancement, respectively.

As summarized in Table~\ref{tab:comparison}, VIRAL achieves state-of-the-art performance across all task categories, outperforming existing baselines by a substantial margin. Most notably, in the interactive segmentation task, our framework yields a two-fold improvement in IoU compared to the strongest competitor. We attribute these performance gaps to the architectural limitations of prior methods when handling heterogeneous tasks. Specifically, while PromptDiff shows reasonable capability in structurally aligned tasks (e.g., Edge detection), its reliance on ControlNet-like spatial conditioning severely hinders its generalization to non-spatial transformations such as deraining or object detection. Conversely, MAE-style inpainting models, such as Painter, IMProv, and VisualPrompt, frequently struggle with fine-grained texture synthesis. These methods tend to produce structural hallucinations or lose high-frequency details, leading to sub-optimal results in colorization and edge detection.

In contrast, by harnessing the generative priors of pre-trained image foundation models and integrating the MoE-LoRA strategy, VIRAL effectively decouples the parameter space for conflicting tasks. This design mitigates gradient interference between geometric perception and generative restoration, allowing the model to maintain high fidelity across the full spectrum of visual tasks. 
Qualitative comparisons are presented in Figure~\ref{fig:tasks_visual}. VIRAL demonstrates remarkable robustness in complex, real-world scenarios where previous approaches often fail to capture the target mapping. It accurately preserves identity and fine-grained details while strictly adhering to the semantic transformation defined by the user-provided examples.

\subsection{Comparison with Task-Specific Methods}
To evaluate the competitive edge of our framework, we benchmark VIRAL against leading domain-specific experts across five representative downstream tasks. Specifically, we compare against SLBR~\cite{Jing2021SLBR} for watermark removal, DepthAnything~\cite{Lihe2024depthany} for depth estimation, StableNormal~\cite{Chongjie2024stalenormal} for surface normal estimation, CSUD~\cite{Guanglu2025csud} for deraining, and CHVI~\cite{Qingsen2025hvi} for low-light enhancement. For a fair comparison, we utilize the official checkpoints of each specialist model, which are fully optimized on their respective benchmark datasets (e.g., using models trained specifically on LOLv2 for low-light enhancement).

The quantitative results, detailed in Table~\ref{tab:reconstruction} and Table~\ref{tab:depth}, indicate that our generalist method achieves performance comparable to, and in many cases significantly surpassing, state-of-the-art specialized models. Regarding generative restoration tasks, VIRAL outperforms the specialist SLBR by a large margin  in watermark removal and achieves superior PSNR in low-light enhancement compared to CHVI. While there is a slight performance gap in the deraining task compared to CSUD, visual inspection reveals that our results remain perceptually indistinguishable from ground truth, prioritizing semantic consistency over pixel-level noise fitting (see Appendix \ref{sec:deraining_analysis}).
In the realm of geometric estimation, VIRAL surpasses both DepthAnything and StableNormal across all metrics. The  ``Ours (single)'' denotes a baseline trained exclusively on the corresponding individual task. The comparison reveals that our unified training strategy successfully circumvents the negative transfer often observed in multi-task learning, exhibiting no performance degradation compared to the single-task counterparts. Details are provided in Appendix~\ref{sec:ablation}.

These empirical results demonstrate that, through the V-ICL paradigm, a pre-trained vision foundation model can attain or even exceed the efficacy of manually engineered, bespoke pipelines across a majority of downstream tasks. We attribute this success primarily to the synergy between the massive world knowledge and high-dimensional generative priors encapsulated in the frozen DiT backbone, and our unified generative formulation of V-ICL, which effectively aligns diverse visual tasks into a coherent inference process.

\begin{table}[t]
\small
\centering
\caption{Quantitative comparison of three Image reconstruction tasks with specialized models.}
\label{tab:reconstruction}
\setlength{\tabcolsep}{12pt}
\begin{tabular}{c|c|cc} 
\toprule
\textbf{Task}                                                                         & \textbf{Method}        & \textbf{PSNR $\uparrow$ }          & \textbf{SSIM $\uparrow$}          \\
\midrule
\multirow{2}{*}{Derain}                                                      & CSUD        & \textbf{33.31}  & \textbf{0.957}  \\
                                                                             & \textbf{Ours} & 29.67           & 0.889           \\
\midrule                                                                            
\multirow{2}{*}{\begin{tabular}[c]{@{}c@{}}Light\\~Enhance\end{tabular}}     & CHVI        & 24.797          & \textbf{0.919}  \\
                                                                             & \textbf{Ours}          & \textbf{25.248} & 0.878           \\
\midrule                                                                             
\multirow{2}{*}{\begin{tabular}[c]{@{}c@{}}Watermark \\Removal\end{tabular}} & SLBR        & 30.972          & 0.934           \\
                                                                             & \textbf{Ours}          & \textbf{36.958} & \textbf{0.959}  \\
\bottomrule
\end{tabular}
\end{table}

\begin{table}[t]
\small
\centering
\caption{Quantitative comparison of surface normal estimation and depth estimation with specialized models.}
\label{tab:depth}
\setlength{\tabcolsep}{9pt}
\begin{tabular}{c|c|cc} 
\toprule
\textbf{Task}                    & \textbf{Method}          & \textbf{AbsRel $\downarrow$}          & \textbf{$\delta_1 \uparrow$ } \\
\midrule
\multirow{3}{*}{Depth}  & DepthAnything & 0.154           & 0.775                  \\
                        & Ours(single)    & 0.145           & 0.796                  \\
                        & \textbf{Ours}   & \textbf{0.138}  & \textbf{0.812}         \\
\midrule
\textbf{Task}                    & \textbf{Method}          & \textbf{Med $\downarrow$  }          & \textbf{Mean $\downarrow$ }                 \\
\midrule
\multirow{3}{*}{Normal} & StableNormal  &  21.352          & 25.845                \\
                        & Ours(single)    &  18.295          & 21.913             \\
                        & \textbf{Ours}   & \textbf{17.582} & \textbf{20.187}       \\
\bottomrule
\end{tabular}
\end{table}

\subsection{Open-Domain In-Context Editing Ability}

\begin{table}[t]
\small
\centering
\caption{Quantitative comparison of general editing task with instruction-based image editing models.}
\label{tab:general}
\setlength{\tabcolsep}{5pt}
\begin{tabular}{c|c|ccc} 
\toprule
\textbf{Task}                                                              & \textbf{Method} & \textbf{CLIP $\uparrow$}  & \textbf{LPIPS $\downarrow$} & \textbf{DINO $\uparrow$}   \\
\midrule
\multirow{3}{*}{\begin{tabular}[c]{@{}c@{}}Open-domain\\~Editing\end{tabular}} & qwen-edit       & 0.845          & 0.652          & 0.734           \\
                                                                           & ICEdit          & 0.823          & 0.621          & 0.717           \\
                                                                           & \textbf{Ours}   & \textbf{0.880} & \textbf{0.517} & \textbf{0.798}  \\
\midrule
\multirow{3}{*}{\begin{tabular}[c]{@{}c@{}}Style \\Transfer\end{tabular}}  & qwen-edit       & 0.701          & 0.814            & 0.598             \\
                                                                           & ICEdit          & 0.710            & 0.813            & 0.653             \\
                                                                           & \textbf{Ours}   & \textbf{0.832} & \textbf{0.718}            & \textbf{0.757}             \\
\bottomrule
\end{tabular}
\end{table}

\begin{figure}[t]
    \centering
    \includegraphics[width=\linewidth]{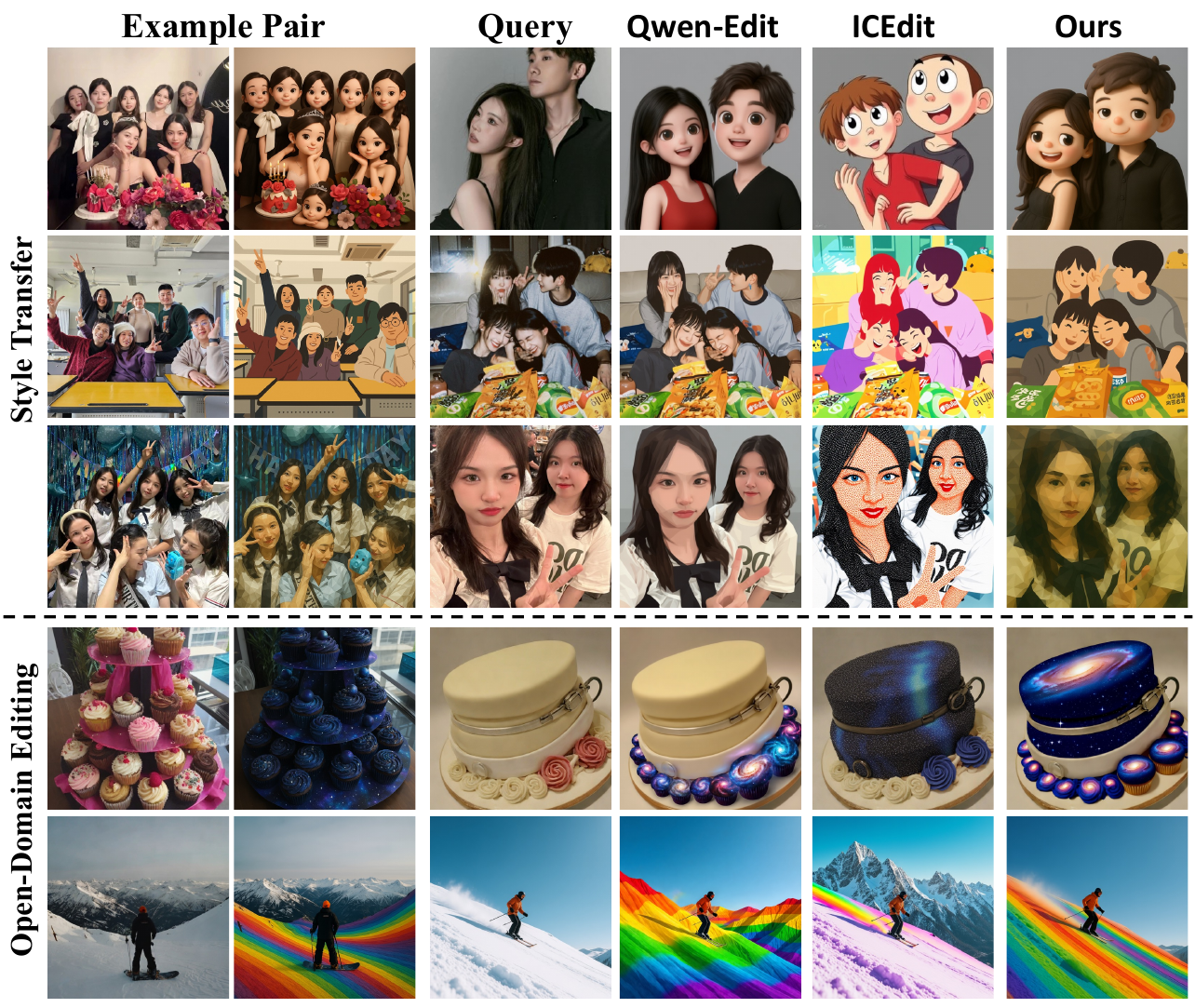}
    \caption{Quantitative comparison on open-domain editing tasks. For style transfer, instruction-driven models often struggle to achieve the desired results. In contrast, our visual in-context demonstrations ensure both stylistic consistency and content preservation. The text editing instructions are provided in Appendix \ref{sec:appendix_prompts}.}
    \label{fig:style}
\end{figure}

We emphasize that the in-context reasoning capability of VIRAL is generalizable beyond specific tasks. We evaluate VIRAL against instruction editing model, including Qwen-Image-Edit and ICEdit~\cite{Zechuan2025ICEdit}. 
To ensure a rigorous and fair comparison, we employ Qwen2.5-VL-7B-Instruct \cite{Qwen2.5-VL} to transcribe the visual demonstrations into precise text instructions for these baselines. 
We specifically focus on Style Transfer, a representative editing task where textual descriptions are often insufficient to accurately capture the visual style. We evaluate performance using 20 diverse styles from OmniConsistency~\cite{Song2025OmniConsistencyLS}. We employ CLIP~\cite{Alec2021clip}, LPIPS~\cite{Richard2018lpips}, and DINO~\cite{Maxime2024dino} to measure the semantic and perceptual similarity between the generated results and the ground truth target.

Table~\ref{tab:general} shows that VIRAL achieves a significant performance leap across all metrics. Specifically, the lower LPIPS indicates better preservation of fine-grained textures, while higher DINO similarity confirms superior object-level semantic consistency. These results demonstrate that VIRAL maintains subject integrity more effectively than text-based image edit models.

These results highlight that textual instructions often suffer from semantic ambiguity, leading to unintended shifts in global layout or stylistic ``hallucinations''. In contrast, visual demonstrations provide dense, unambiguous pixel-level guidance. This advantage is particularly pronounced in style transfer; as this task is inherently difficult to articulate linguistically, our visual conditioning provides rich, non-parametric semantic information that facilitates precise analogy-making.
Moreover, our method outperforms its backbone, Qwen-Image-Edit. This advancement suggests that our in-context fine-tuning stage does not merely exploit existing capabilities but effectively elicits dormant reasoning potentials within the pre-trained DiT. Qualitative visualizations in Figure~\ref{fig:style} further substantiate these findings, showcasing our model's fidelity in executing diverse, high-level semantic edits.
We provide more experiments on model generalization and robustness in Appendix \ref{sec:generalization_study} and \ref{sec:ablation}.

\section{Conclusion}

% In this work, we present a unified framework that elicits visual in-context reasoning capabilities from pre-trained image editing models, eliminating the need to train task-specific learners from scratch. By formulating Visual In-Context Learning (V-ICL) as a visual analogy conditional generation problem, our approach seamlessly integrates diverse tasks—ranging from low-level image restoration to high-level semantic editing—into a single RGB space. We demonstrate that by combining a frozen Diffusion Transformer (DiT) backbone with role-aware token conditioning and parameter-efficient fine-tuning, we can perform a wide range of in-context editing tasks while preserving the model’s vast generative priors. Furthermore, to facilitate this paradigm shift, we construct and release a comprehensive In-Context Editing Dataset spanning standard perception, image restoration, and open-domain instruction-based editing. Extensive experiments show that our model significantly outperforms existing V-ICL baselines and achieves competitive performance against specialized domain experts. We hope this work inspires further research into building universal visual generalists that can flexibly adapt to user needs through simple visual demonstrations.

In this work, we present a unified framework that elicits visual in-context reasoning capabilities from pre-trained image editing models. This approach eliminates the necessity of training task-specific learners from scratch. By formulating Visual In-Context Learning (V-ICL) as a visual analogy conditional generation problem, our framework integrates diverse tasks into a single RGB space. These tasks encompass a broad spectrum from low-level image restoration to high-level semantic editing. We demonstrate that combining a frozen DiT backbone with role-aware token conditioning and parameter-efficient fine-tuning enables a wide range of in-context editing tasks. Importantly, this strategy preserves the model’s extensive generative priors. Furthermore, we introduce a comprehensive In-Context Editing Dataset spanning standard perception, image restoration, and open-domain instruction-based editing to facilitate this paradigm shift. Extensive experiments confirm that our model significantly outperforms existing V-ICL baselines and achieves competitive performance against specialized domain experts. We hope this work inspires further research into building universal visual generalists that flexibly adapt to user needs through visual demonstrations.
% \section{Electronic Submission}

\section*{Impact Statement}

% Authors are \textbf{required} to include a statement of the potential broader
% impact of their work, including its ethical aspects and future societal
% consequences. This statement should be in an unnumbered section at the end of
% the paper (co-located with Acknowledgements -- the two may appear in either
% order, but both must be before References), and does not count toward the paper
% page limit. In many cases, where the ethical impacts and expected societal
% implications are those that are well established when advancing the field of
% Machine Learning, substantial discussion is not required, and a simple
% statement such as the following will suffice:

This paper presents work whose goal is to advance the field of Machine
Learning. There are many potential societal consequences of our work, none
which we feel must be specifically highlighted here.

% The above statement can be used verbatim in such cases, but we encourage
% authors to think about whether there is content which does warrant further
% discussion, as this statement will be apparent if the paper is later flagged
% for ethics review.

\nocite{langley00}

\bibliography{ref}
\bibliographystyle{icml2026}

%%%%%%%%%%%%%%%%%%%%%%%%%%%%%%%%%%%%%%%%%%%%%%%%%%%%%%%%%%%%%%%%%%%%%%%%%%%%%%%
%%%%%%%%%%%%%%%%%%%%%%%%%%%%%%%%%%%%%%%%%%%%%%%%%%%%%%%%%%%%%%%%%%%%%%%%%%%%%%%
% APPENDIX
%%%%%%%%%%%%%%%%%%%%%%%%%%%%%%%%%%%%%%%%%%%%%%%%%%%%%%%%%%%%%%%%%%%%%%%%%%%%%%%
%%%%%%%%%%%%%%%%%%%%%%%%%%%%%%%%%%%%%%%%%%%%%%%%%%%%%%%%%%%%%%%%%%%%%%%%%%%%%%%

\newpage
\appendix
\onecolumn
\section*{Appendix Contents}
\begin{itemize}
    \setlength{\itemsep}{0.5em}
    
    \item \textbf{Appendix A: Implementation Details} \dotfill Page \pageref{sec:appendix_impl}
        \begin{itemize}
            \setlength{\itemsep}{0.2em}
            \item[--] A.1 Detailed Implementation of VIRAL \dotfill Page \pageref{sec:a1}
            \item[--] A.2 Comparison Baselines \dotfill Page \pageref{sec:a2}
            \item[--] A.3 Hyperparameters for Dataset Construction \dotfill Page \pageref{sec:appendix_dataset_params}
            \item[--] A.4 LLM Prompts for Data Synthesis \dotfill Page \pageref{sec:a4}
        \end{itemize}

    \item \textbf{Appendix B: Generalization Study} \dotfill Page \pageref{sec:generalization_study}

    \item \textbf{Appendix C: Ablation Study} \dotfill Page \pageref{sec:ablation}
        \begin{itemize}
            \setlength{\itemsep}{0.2em}
            \item[--] C.1 Single-task training and joint training \dotfill Page \pageref{sec:c1}
            \item[--] C.2 Model designs \dotfill Page \pageref{sec:c2}
            \item[--] C.3 Robustness to Exemplar Selection \dotfill Page \pageref{sec:c3}
            \item[--] C.4 Cross-Task Generalization and Domain Robustness \dotfill Page \pageref{sec:c4}
        \end{itemize}

    \item \textbf{Appendix D: Bidirectional Visual Translation} \dotfill Page \pageref{sec:inver}

    \item \textbf{Appendix E: Qualitative Analysis on Image Deraining} \dotfill Page \pageref{sec:deraining_analysis}

    \item \textbf{Appendix F: Visualization of the In-Context Dataset} \dotfill Page \pageref{sec:dataset_sample}

    \item \textbf{Appendix G: More visualizations} \dotfill Page \pageref{sec:more_vis}

    \item \textbf{Appendix H: Detailed Editing Instructions} \dotfill Page \pageref{sec:appendix_prompts}
        
\end{itemize}

\section{Implementation Details}
\label{sec:appendix_impl}
\subsection{Detailed Implementation of VIRAL}
\label{sec:a1}
All experiments are conducted based on the \texttt{Qwen-Image-Edit-2511} architecture~\cite{wu2025qwenimagetechnicalreport}. Below, we detail the specific configurations for model adaptation, training strategy, and inference protocols.

\paragraph{Model Configuration and Hybrid Adaptation.}
The model is initialized from the official pre-trained weights. To achieve parameter-efficient adaptation while handling task heterogeneity, we target the attention mechanisms (Query, Key, Value), output projections (\texttt{to\_out}), and Feed-Forward Networks (FFN) across all DiT blocks. We employ a hybrid adaptation strategy:
\begin{itemize}[leftmargin=*, topsep=0pt, itemsep=0pt]
    \item \textbf{MoE-LoRA:} Integrated exclusively at the output projection layer of the FFNs. We configure this with $N=4$ experts, a rank of $r=64$, and utilize Top-2 routing.
    \item \textbf{Standard LoRA:} Applied to all other targeted layers (e.g., Attention $Q/K/V$) with a rank of $r=64$.
\end{itemize}

\paragraph{Training Dynamics.}
Our unified multi-task training is conducted on 8 NVIDIA A800 GPUs for approximately 48 hours.
\begin{itemize}[leftmargin=*, topsep=0pt, itemsep=0pt]
    \item \textbf{Optimization:} We use the AdamW optimizer with a constant learning rate of $1 \times 10^{-4}$ and a per-device batch size of 1.
    \item \textbf{Decoupled Task Sampling:} To ensure training stability, we employ a decoupled sampling strategy. At each iteration, each GPU independently selects a task category and samples a data pair. Consequently, the effective global batch across the 8 devices is composed of a diverse mixture of tasks, preventing gradient conflict and overfitting to specific modalities.
\end{itemize}

\paragraph{Inference Protocol.}
During inference, we denoise 40 steps, strictly following the base model's default configuration.
To rigorously evaluate In-Context Learning (ICL) performance:
\begin{itemize}[leftmargin=*, topsep=0pt, itemsep=0pt]
    \item \textbf{Context Setup:} We strictly adhere to a 1-shot setting. For standard visual tasks, the model is provided with a single visual context pair randomly selected from the \textbf{held-out test set} of the corresponding task.
    \item \textbf{Evaluation:} All quantitative metrics are computed on this pre-defined test split to ensure no data leakage.
\end{itemize}

\subsection{Comparison Baselines}
\label{sec:a2}
For V-ICL baselines—specifically VisualPrompt, IMprov, Painter, and PromptDiffusion—we utilize their officially released checkpoints and adhere to the hyperparameter configurations provided in their respective original papers. 
To ensure optimal performance for multi-modal baselines capable of processing text, we provide task-specific textual instructions alongside visual inputs. For instance, for IMprov, we employ its standard prompting format (e.g., ``Left-input image, right-depth/surface normal estimation''), and for PromptDiffusion, we supply the corresponding task descriptor (e.g., ``depth map'').

To accommodate the varying resolution constraints of each architecture, input images are resized to match their native requirements while maintaining fair comparison standards. 
Specifically, for models operating on a $2 \times 2$ grid layout (VisualPrompt and IMprov) with a total resolution of $224 \times 224$, individual images are resized to $112 \times 112$ before concatenation. 
Similarly, for Painter, which supports a resolution of $448 \times 448$, individual images are resized to $224 \times 224$. 
PromptDiffusion is evaluated at its native input resolution of $512 \times 512$.
Following inference, architecture-specific post-processing is applied to decode outputs into standard RGB formats. Finally, all predictions are resized to the original ground-truth resolution to ensure standardized metric calculation.

Crucially, to enforce strict comparability, we fix the random seed and the selected exemplar pair for every test query across all methods (including our VIRAL framework). These exemplar pairs are randomly sampled from the held-out test set.

\subsection{Hyperparameters for Dataset Construction}
\label{sec:appendix_dataset_params}
Here, we provide the precise hyperparameters used in our data curation process.

\begin{itemize}[leftmargin=*]
    \item \textbf{Adaptive Clustering Setup:} For the unsupervised organization of pre-trained CLIP vectors, we determine the number of clusters $K$ based on the scale of each dataset subset. Let $N$ denote the number of samples in a subset; we adopt a dynamic assignment strategy:
    \begin{equation}
        K = 
        \begin{cases} 
        1500 & \text{if } N < 25,000 \\
        3000 & \text{if } N \ge 25,000 
        \end{cases}
    \end{equation}
    This ensures sufficient granularity for large-scale subsets while preventing over-segmentation in smaller ones.
    
    \item \textbf{Dual Filtering Strategy:} To ensure data quality, we enforced strict numerical thresholds:
    \begin{enumerate}
        \item \textbf{Visual De-duplication:} To eliminate visually redundant source images, we discarded pairs with a visual similarity score higher than $\tau_{vis} = 0.98$ (calculated via Cosine Similarity).
        \item \textbf{Textual Alignment:} To guarantee high semantic consistency between the visual content and instructions, we enforced a minimum text-image similarity threshold of $\tau_{text} > 0.9$.
    \end{enumerate}
\end{itemize}

\subsection{LLM Prompts for Data Synthesis}
\label{sec:a4}
To facilitate \textbf{Generative Analogous Synthesis}, we employ the Qwen-Max API to generate descriptive captions for novel scenes. The objective is to synthesize a semantically distinct source description that maintains logical compatibility with the original editing instruction. The specific system prompt provided to the LLM is detailed below:

\begin{tcolorbox}[colback=gray!10, colframe=gray!50, title=\textbf{System Prompt for Analogous Caption Generation}]
\small
The user will input an image description and an editing command.

Your task is to replace as many entities or attributes as possible in the image description to generate a new image description. However, the editing command must remain valid for the rewritten image.

You should first analyze the editing command to determine what modifications it made. Ensure that the rewritten image description can also be edited using this command. For example, if the command changes A to B, the rewritten description should explicitly include A. If the command adds A, then A should not appear in the rewritten description. Entities not mentioned in the instructions can be freely replaced with other entities.

\textbf{Examples:}
\begin{itemize}[leftmargin=*]
    \item \textbf{Image Description:} ``A sleek black sports car with a closed top parked on the side of a mountain road at sunset.''
    \item \textbf{Editing Instruction:} ``Change the color of the sports car to red and make it a convertible.''
    \item \textbf{Acceptable Rewrite:} ``A glossy black coupe parked along a coastal cliff road during golden hour.''
\end{itemize}
\textit{(Rationale: The scene remains a non-red, non-convertible sports car parked in a sunset-like setting—thus editable. Since the instruction modifies the car's color and type, the rewritten description must include the car but preserve its original attributes. The background, not targeted by the instruction, is altered from a mountain road to a coastal cliff.)}

\textbf{Note:} The rewritten image description should be clearly different from the original input image description, but it shouldn't describe the edited scene. The rewritten caption cannot be the edited scene.

The rewritten description should be enclosed in \texttt{<rw></rw>} tags.
\end{tcolorbox}

For the \textbf{Open-Domain Editing} component, we utilize Qwen2.5-VL-7B-Instruct to derive textual editing instructions directly from visual exemplar pairs ($x_s, s_t$). The model is prompted with the following directive:

\begin{tcolorbox}[colback=gray!10, colframe=gray!50, title=\textbf{Prompt for Instruction Generation (Qwen2.5-VL)}]
\small
``Compare Image A and Image B. Write a concise English editing instruction to transform Image A into Image B. Only return the string of the editing instructions; nothing else.''
\end{tcolorbox}

\section{Generalization Study}
\label{sec:generalization_study}

To rigorously evaluate the robustness and generalization capabilities of our framework, we conduct tests across two challenging dimensions: Out-of-Distribution (OOD) Dataset and Unseen Task Generalization.

We first evaluate the model's robustness to domain shift by testing our object detection performance on the Pascal-5i dataset~\cite{Amirreza2017pascal}, which contains categories and environments distinct from our training corpus. As reported in Table~\ref{tab:generalization}, our model maintains high localization accuracy and semantic alignment despite the distributional shift. This performance indicates that the model has internalized the logic of object detection via visual analogy, rather than merely memorizing training-specific image patterns.

\begin{figure}
    \centering
    \includegraphics[width=0.99\linewidth]{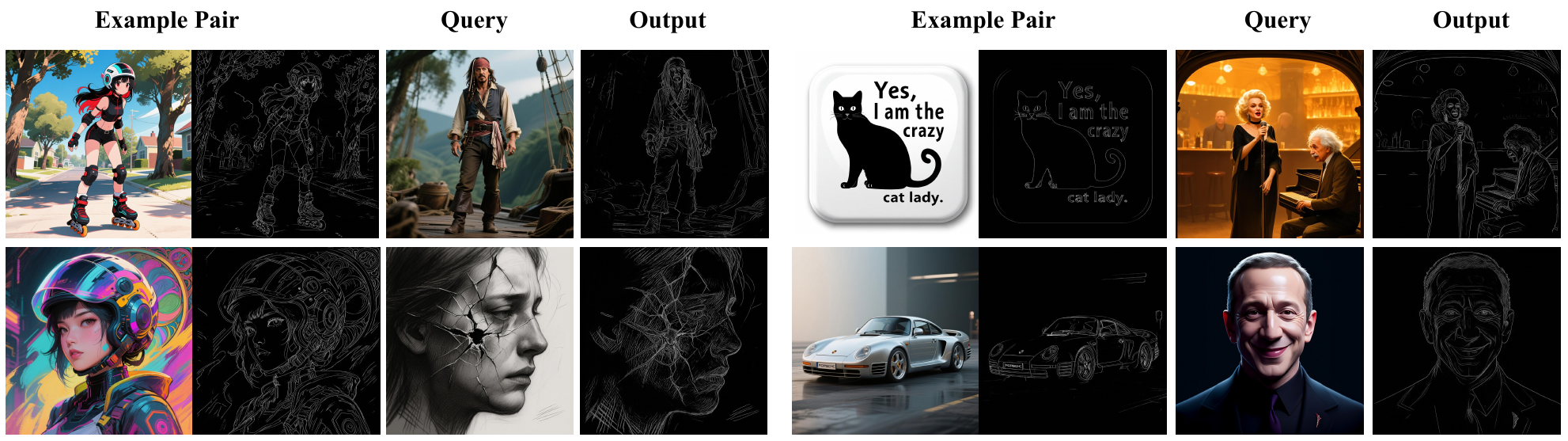}
    \caption{Zero-shot generalization to unseen Lineart Generation.
Despite being trained exclusively on standard Canny edge maps, VIRAL successfully generalizes to the artistic lineart task via one-shot in-context learning. }
    \label{fig:lineart_gen}
\end{figure}

To further stress-test the abstract reasoning capabilities of VIRAL, we evaluate it on Lineart Generation, a task strictly excluded from our training phase. 
It is crucial to distinguish this from the standard Edge Detection  seen during training. unlike edge detection which relies on low-level pixel gradients, Lineart requires a higher-level semantic abstraction to render clean, artistic contours.
As qualitatively demonstrated in Figure~\ref{fig:lineart_gen}, despite never observing lineart data during training, our model successfully infers this specific artistic style from a single exemplar pair $(x_s, x_t)$ and generalizes the transformation to unseen query images.
This capability is of significant practical value, allowing end-users to deploy the model for customized, niche tasks without requiring specialized fine-tuning.
\begin{table}[ht]
\small
\centering
\caption{Quantitative comparison of object detection and lineart estimation.}
\label{tab:generalization}
\begin{tabular}{c|c|cc} 
\toprule

\multirow{2}{*}{\textbf{Method}} & \textbf{Object Det.} & \multicolumn{2}{c}{\textbf{Lineart}}  \\

                        & IoU $\uparrow$               & RMSE $\downarrow$ & LPIPS $\downarrow$                \\
\midrule
IMProv                  & 0.251              & 80.25     &  0.726                     \\
VisualPrompt            & 0.318              & 50.81     & 0.645                      \\
PromptDiff              & 0.326              & 189.89     & 0.780                      \\
Painter                 & 0.458              & 82.22     & 0.650                      \\
\textbf{Ours}           & \textbf{0.721}     & \textbf{34.82}     & \textbf{0.339}                      \\
\bottomrule
\end{tabular}
\end{table}

\section{Ablation Study}
\label{sec:ablation}
\subsection{Single-task training and joint training} 
\label{sec:c1}
To investigate the potential synergistic or competitive effects between heterogeneous tasks, we conduct a comparative analysis between multi-task joint training and single-task optimization. We specifically isolate depth estimation and surface normal estimation for this study. To ensure a controlled comparison, we maintain identical total training iterations, data volume per task, and hyperparameter configurations for both settings. The results, denoted as ``Ours (Single)'' in Table~\ref{tab:depth}, reveal that joint training achieves marginal performance gains over single-task models. While the improvement is not transformative, the absence of performance degradation (i.e., negative transfer) is significant. This suggests that our framework, bolstered by the MoE-LoRA architecture, effectively mitigates inter-task interference and successfully aggregates geometric priors across different domains. We hypothesize that the current performance reflects a state of performance saturation within the scope of the current dataset scale. It is highly probable that as the diversity and volume of the In-Context dataset further scale up, the advantages of joint training in fostering cross-task transfer and visual reasoning will become more pronounced.

\subsection{Model designs}
\label{sec:c2}
% Effectiveness of MoE-LoRA. 
To validate our architectural design, we conduct a quantitative comparison between the proposed MoE-LoRA and a standard LoRA baseline. As reported in Table~\ref{tab:lora}, MoE-LoRA consistently outperforms the single-adapter counterpart across all evaluated tasks. Notably, we observe a substantial improvement in segmentation IoU ($+8.1\%$), suggesting that the mixture-of-experts mechanism is particularly effective at handling high-level semantic variations. Crucially, this performance gain is achieved with a modest parameter increase of approximately 10\%, a efficiency attributed to our strategic design where MoE modules are applied exclusively to the output projection layer of the Feed-Forward Network (FFN). These results confirm that MoE-LoRA effectively mitigates gradient interference arising from task heterogeneity without imposing a heavy computational burden.
\begin{table}[ht]
\small
\centering
\caption{Ablation study on adapter architectures. MoE-LoRA consistently outperforms the standard LoRA baseline across all evaluated tasks.}
\label{tab:lora}
\setlength{\tabcolsep}{2pt}
\begin{tabular}{c|ccccc} 
\toprule
\multirow{2}{*}{Method} & \multicolumn{2}{c}{Depth} & \multicolumn{2}{c}{Normal} & Segmentation \\

 & AbsRel $\downarrow$ & $\delta_1$ $\uparrow$ & Med $\downarrow$ & Mean $\downarrow$ & IoU $\uparrow$ \\
\midrule
Standard LoRA & 0.143 & 0.803 & 17.580 & 21.127 & 0.714 \\
MoE-LoRA & \textbf{0.138} & \textbf{0.812} & \textbf{17.582} & \textbf{20.187} & \textbf{0.795} \\
\bottomrule
\end{tabular}
\end{table}

\subsection{Robustness to Exemplar Selection}
\label{sec:c3}

To evaluate stability against the inherent sensitivity of ICL, we compare a Curated Exemplar with Random Exemplars, where we report the mean and standard deviation derived from 5 randomly sampled pairs. As shown in Table~\ref{tab:robustness}, VIRAL exhibits minimal variance; the curated exemplar performs on par with the random average across tasks. This consistency indicates that our model successfully decouples transformation logic from incidental visual content, performing robust semantic analogy rather than relying on spurious pixel-level alignments. Consequently, VIRAL remains reliable even when user-provided demonstrations are varied or sub-optimal.
\begin{table}[ht]
\small
\centering
\caption{\textbf{Robustness to exemplar selection.} The minimal standard deviation between fixed and random exemplars confirms VIRAL's invariance to specific demonstrations.}
\label{tab:robustness}
\begin{tabular}{c|c|cc} 
\toprule
Task                    & Metric                & Fix Exemplar & Random Exemplars                  \\
\midrule
\multirow{2}{*}{Depth}  & AbsRel $\downarrow$               & 0.1408       & 0.1409 $\pm$ 0.0004  \\
                        & $\delta_1 \uparrow$ & 0.7944       & 0.8085 $\pm$ 0.0012  \\
\midrule
\multirow{2}{*}{Normal} & Med $\downarrow$                   & 20.643       & 20.925 $\pm$ 0.0496 \\
                        & Mean $\downarrow$                 & 18.028      &  18.195 $\pm$ 0.0202  \\
\bottomrule
\end{tabular}
\end{table}

\subsection{Cross-Task Generalization and Domain Robustness}
\label{sec:c4}
\begin{figure*}[t]
    \centering
    \includegraphics[width=0.9\linewidth]{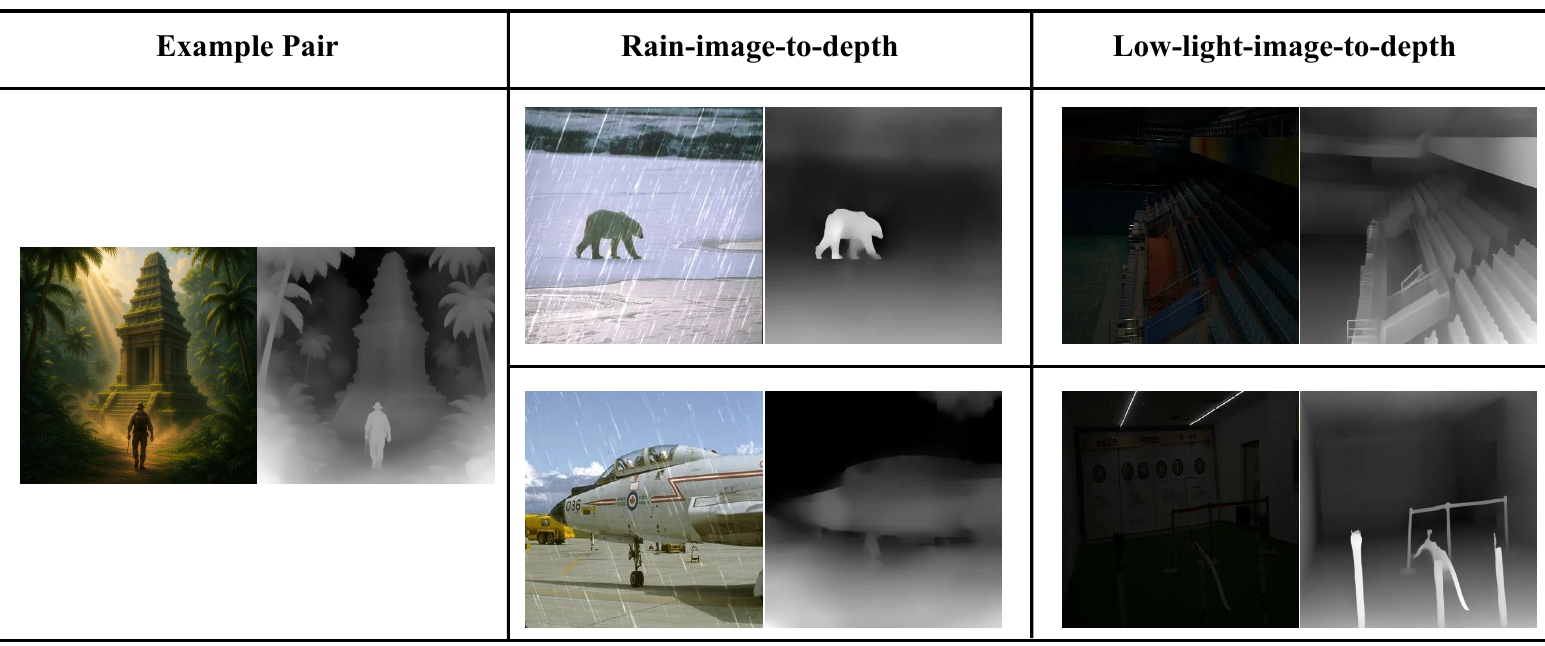}
    \caption{Cross-task generalization and domain robustness. When provided with a \textit{Depth Estimation} exemplar, our model correctly extracts the depth map from query images belonging to unrelated domains (Deraining and Low-light Enhancement). This demonstrates that the visual prompt effectively overrides the inductive biases associated with the query's degradation features.}
    \label{fig:crosstask}
\end{figure*}

To investigate whether visual context effectively dictates task semantics against strong data-driven priors, we conduct a cross-domain inference experiment. We pair a Depth Estimation exemplar with query images sampled from distinct domains, specifically Deraining and Low-light Enhancement. As illustrated in Figure~\ref{fig:crosstask}, VIRAL consistently executes the context-specified depth estimation rather than triggering the restoration tasks typically associated with such degraded inputs. The high fidelity of the generated depth maps demonstrates an emergent capability to perceive underlying geometry amidst severe visual corruption. These results indicate that the exemplar pair functions as a dominant non-parametric instruction, effectively overriding the inductive biases of the query domain. Furthermore, this demonstrates that our model successfully disentangles high-level task logic from low-level image statistics, establishing the framework as a genuine in-context reasoner capable of robust generalization in out-of-distribution scenarios.

\section{Bidirectional Visual Translation}
\label{sec:inver}
\begin{figure*}[t!]
    \centering
    \includegraphics[width=0.99\linewidth]{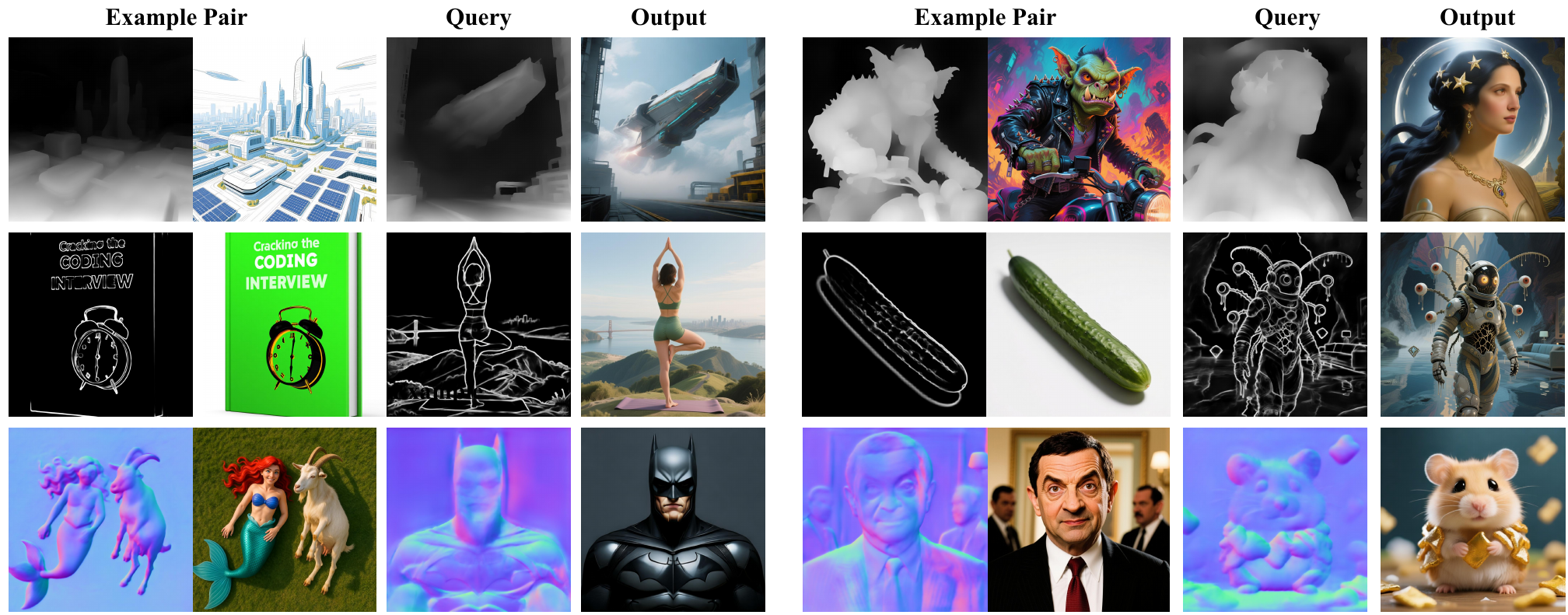}
    \caption{
Qualitative results of inverse tasks. 
VIRAL effectively synthesizes photorealistic RGB images from diverse geometric conditions. 
This figure demonstrate the model's capability to generate high-quality images conditioned on Edges, Depth Maps, and Surface Normals, respectively, while strictly adhering to the provided structural layouts.
}
\label{fig:inverse_tasks}
\end{figure*}

To achieve a holistic understanding of visual scenes, our unified training paradigm explicitly incorporates inverse task learning. Unlike traditional methods that train separate models for perception (RGB $\to$ X) and generation (X $\to$ RGB), VIRAL learns these bidirectional flows simultaneously within a shared parameter space.
As shown in Figure~\ref{fig:inverse_tasks}, this strategy empowers the model to reconstruct high-fidelity photorealistic images from various geometric conditions, including Edge, Depth, and Surface Normal maps, demonstrating its versatility as a universal conditional renderer.

\section{Qualitative Analysis on Image Deraining}
\label{sec:deraining_analysis}
\begin{figure*}[t!]
    \centering
    \includegraphics[width=0.99\linewidth]{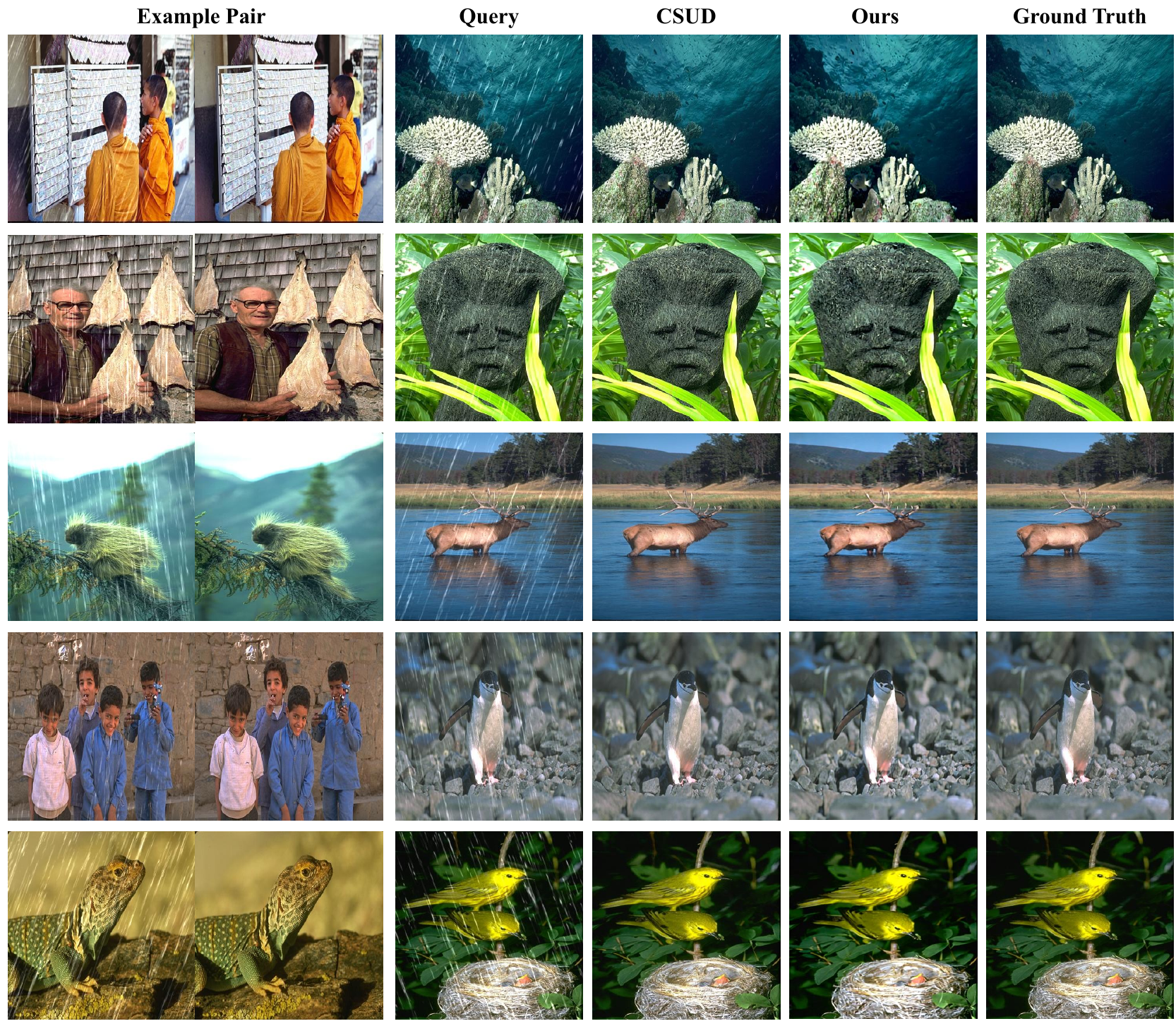}
    \caption{
\textbf{Qualitative comparison of image deraining results.} 
We compare VIRAL against the state-of-the-art specialist CSUD. 
Despite the slight difference in standard metrics, our generative approach achieves effective rain removal with high fidelity, producing images that are visually coherent and nearly identical to the ground truth.
}
\label{fig:deraining_vis}
\end{figure*}

While our method exhibits a marginal numerical gap compared to the specialist model CSUD~\cite{Guanglu2025csud} in quantitative metrics, the visual inspection tells a different story. 
As illustrated in Figure~\ref{fig:deraining_vis}, VIRAL effectively eliminates rain streaks while preserving high-frequency background details, yielding results that are perceptually indistinguishable from the ground truth and prioritizing semantic consistency over pixel-level fitting.

\section{Visualization of the In-Context Dataset}
\label{sec:dataset_sample}
\begin{figure*}[ht]
    \centering
    \includegraphics[width=0.99\linewidth]{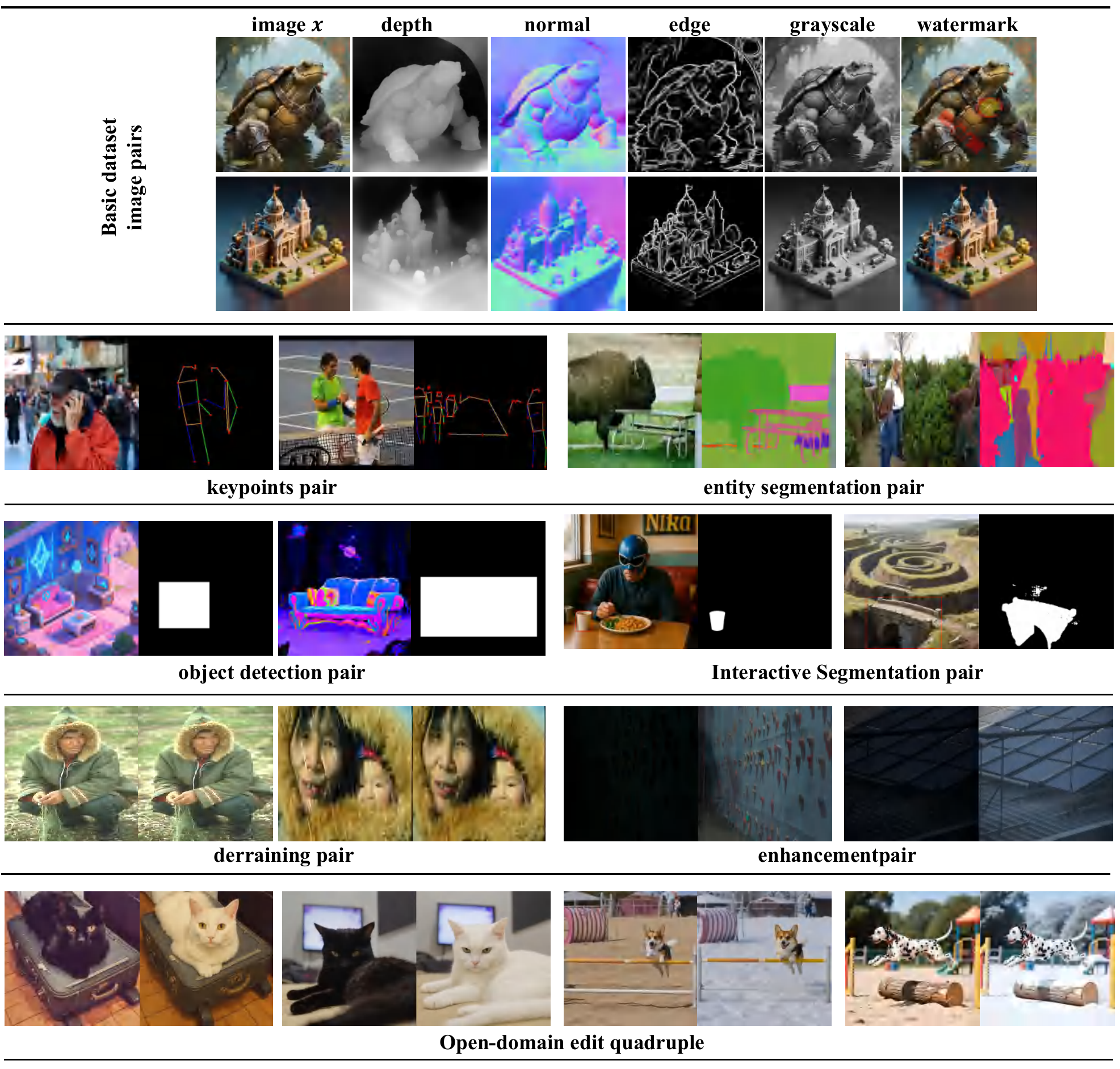}
    \caption{Representative samples from our In-Context Dataset.
The figure displays a diverse collection of visual tasks included in our training corpus. 
}
\label{fig:dataset_samples}
\end{figure*}

To provide a tangible view of our data construction, we visualize representative samples from our proposed In-Context Dataset in Figure~\ref{fig:dataset_samples}. 
These examples highlight the extensive coverage of our corpus, spanning a wide spectrum of visual tasks ranging from fundamental perception and restoration to open-domain creative editing.

\section{More visualizations}
\label{sec:more_vis}
\begin{figure}[t]
    \centering
    \includegraphics[width=0.99\linewidth]{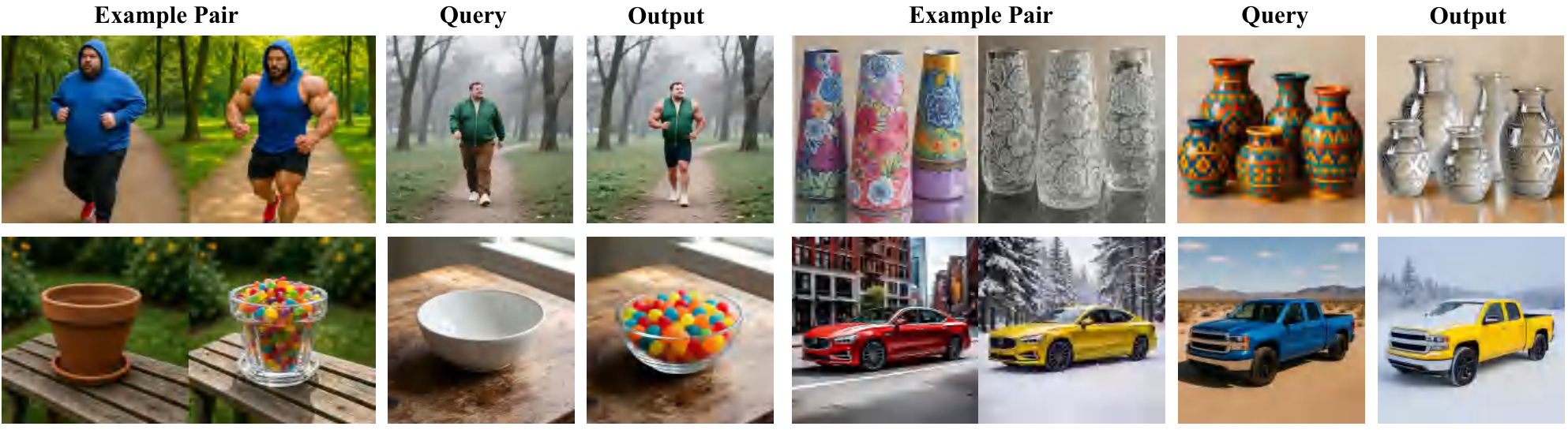}
    \caption{Visualization of open-domain editing.}
    \label{fig:edit_v}
\end{figure}

\begin{figure}[t]
    \centering
    \includegraphics[width=0.99\linewidth]{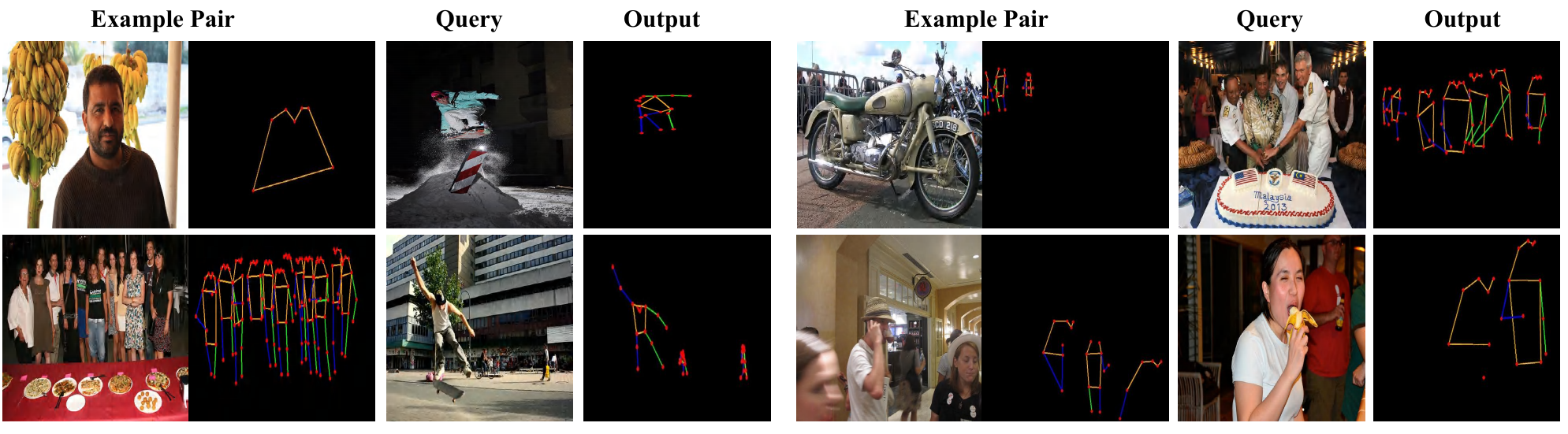}
    \caption{Visualization of human keypoints estimation.}
    \label{fig:1point_v}
\end{figure}

\begin{figure}[t]
    \centering
    \includegraphics[width=0.99\linewidth]{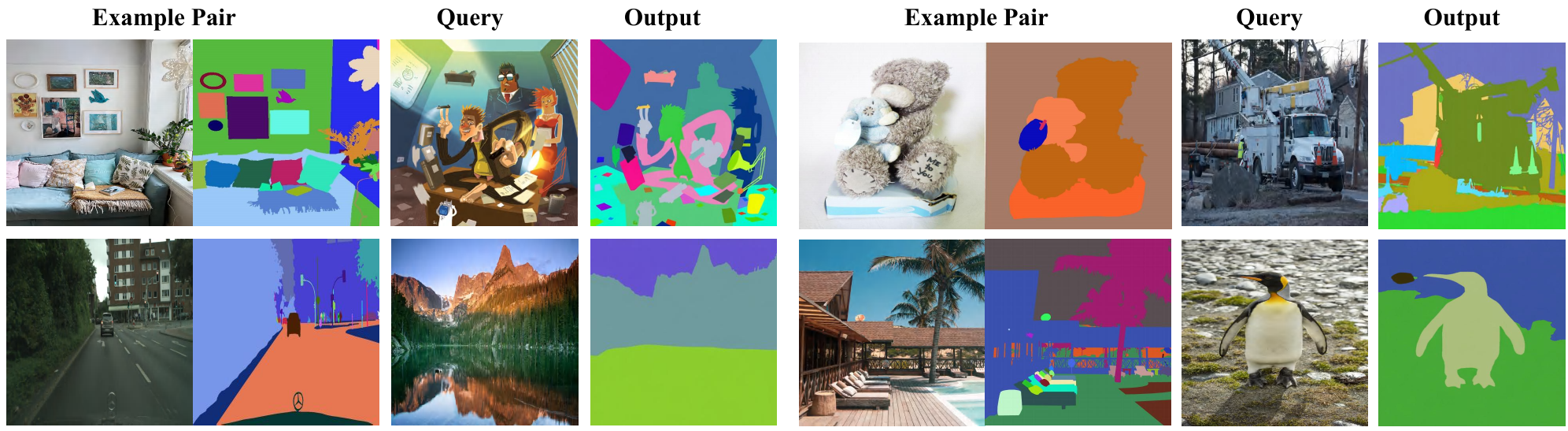}
    \caption{Visualization of entity segmentation.}
    \label{fig:1enseg_v}
\end{figure}

\begin{figure}[t]
    \centering
    \includegraphics[width=0.99\linewidth]{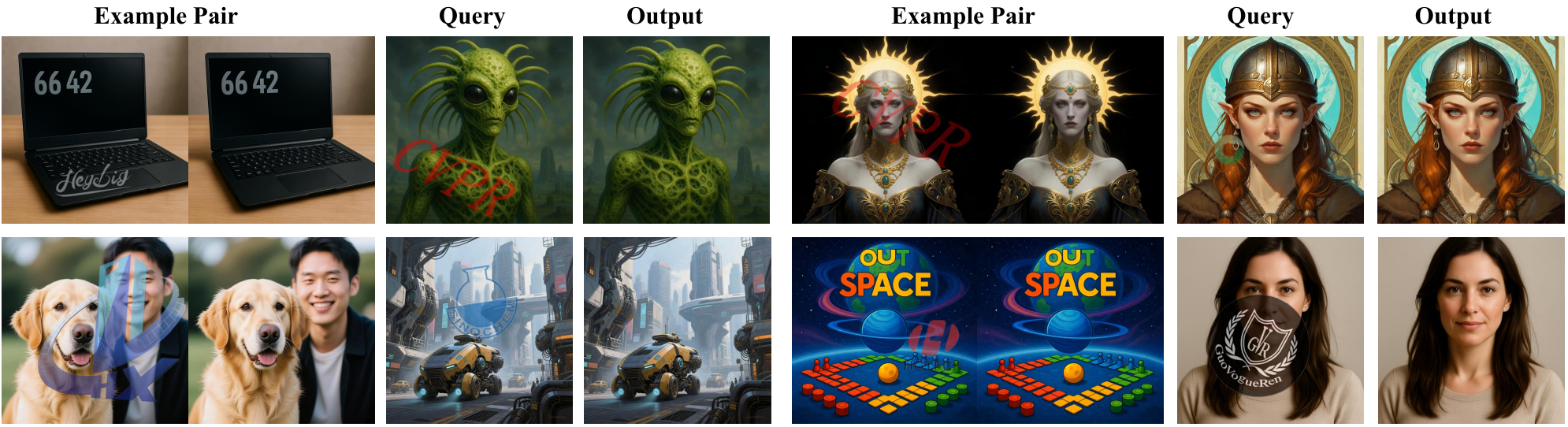}
    \caption{Visualization of watermark removal.}
    \label{fig:1water_v}
\end{figure}

\begin{figure}[t]
    \centering
    \includegraphics[width=0.99\linewidth]{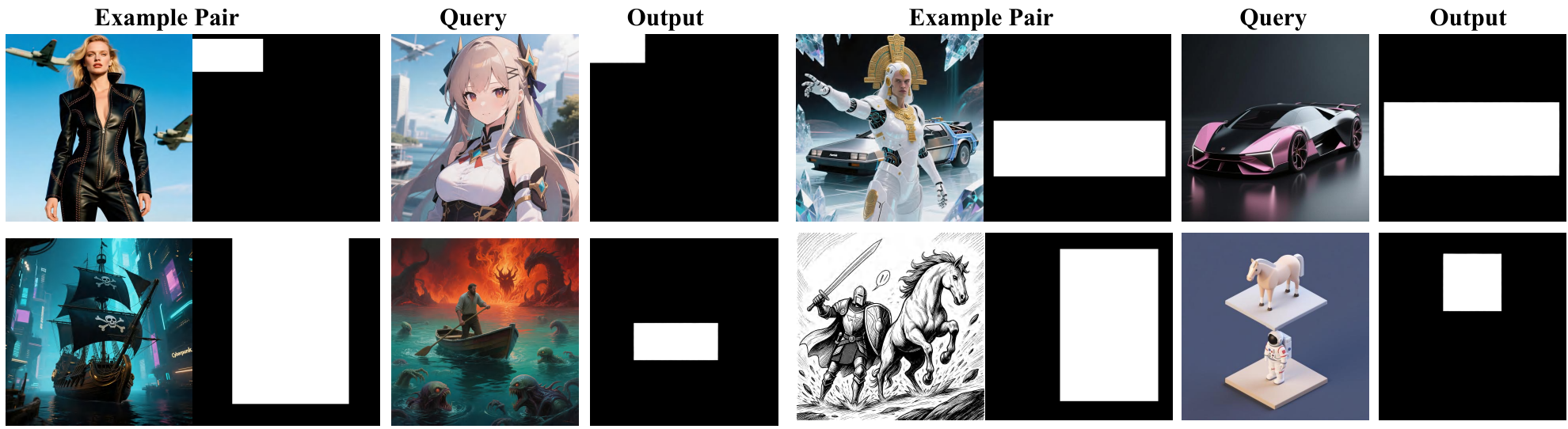}
    \caption{Visualization of object detection.}
    \label{fig:1detecttion}
\end{figure}

\begin{figure}[t]
    \centering
    \includegraphics[width=0.99\linewidth]{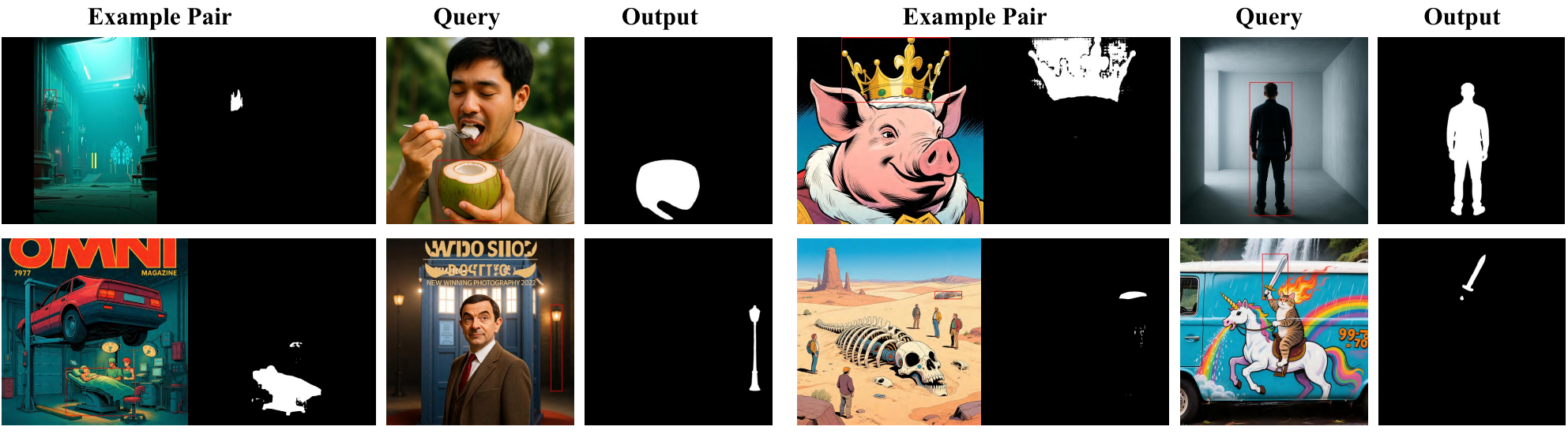}
    \caption{Visualization of interactive segmentation.}
    \label{fig:1seg_v}
\end{figure}

\begin{figure}[t]
    \centering
    \includegraphics[width=0.99\linewidth]{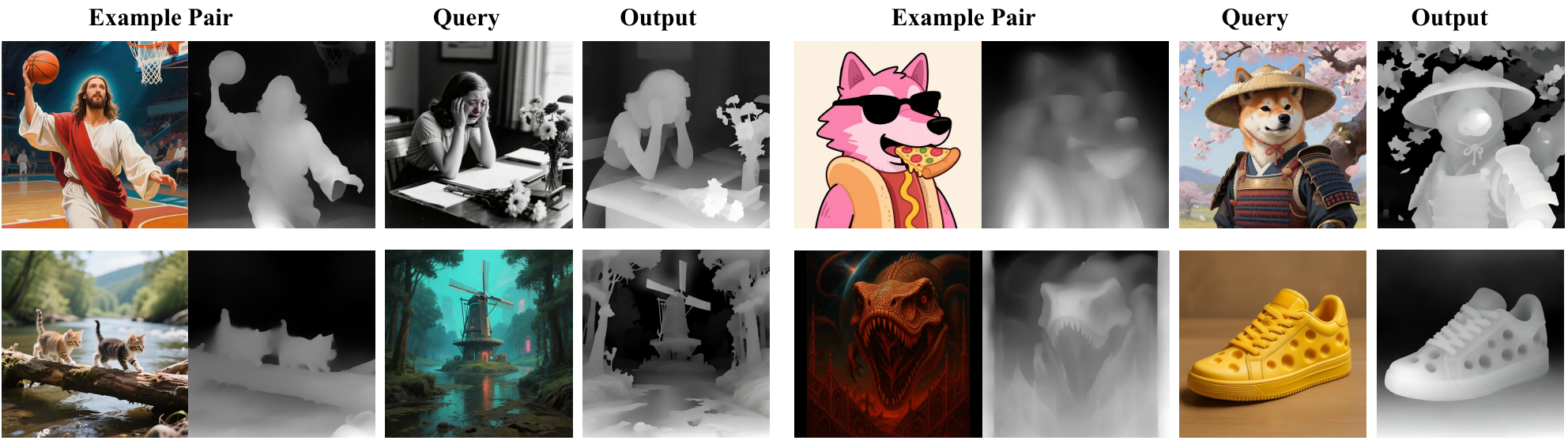}
    \caption{Visualization of depth estimation.}
    \label{fig:1depth_v}
\end{figure}

\begin{figure}[t]
    \centering
    \includegraphics[width=0.99\linewidth]{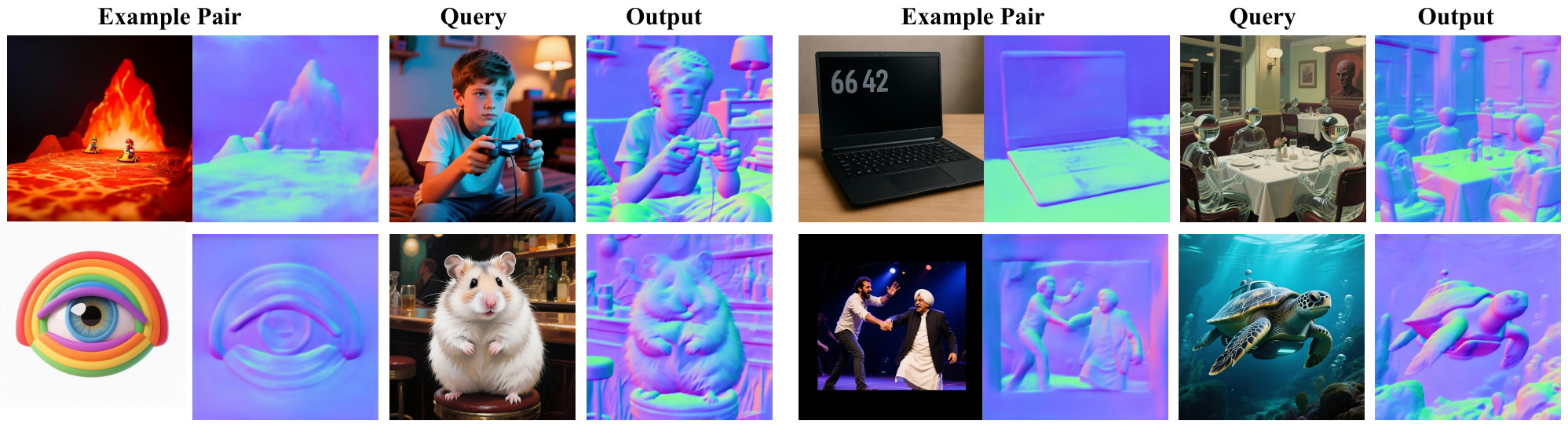}
    \caption{Visualization of Surface normal estimation.}
    \label{fig:1normal_v}
\end{figure}

\begin{figure}[t!]
    \centering
    \includegraphics[width=0.99\linewidth]{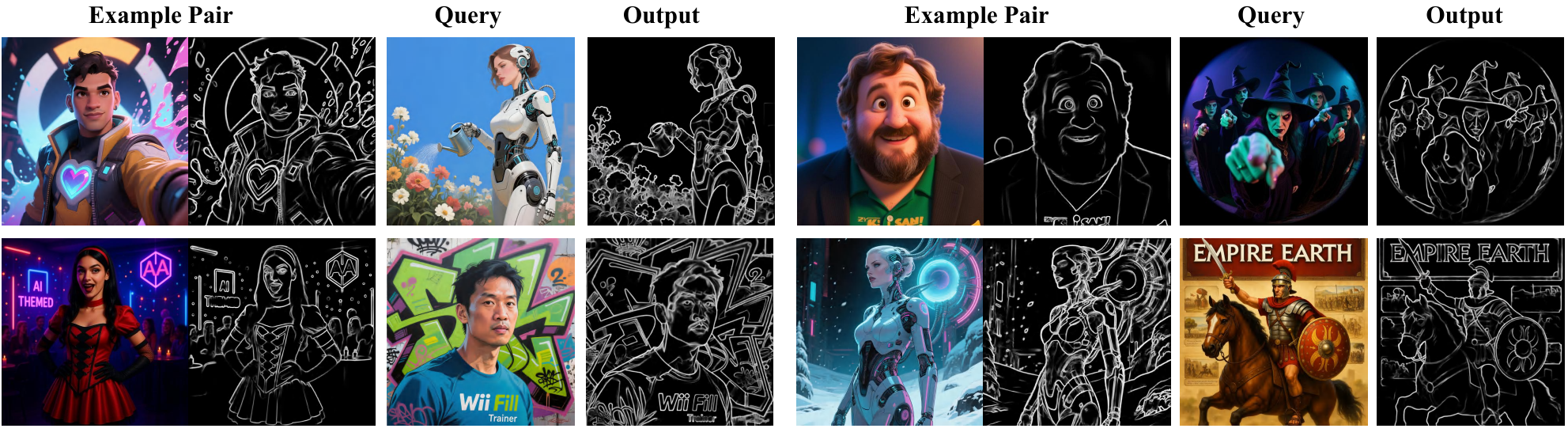}
    \caption{Visualization of edge detection.}
    \label{fig:1softedge_v}
\end{figure}

Here we provide more visualizations of VIRAL for various tasks, as shown in Figure \ref{fig:edit_v}, \ref{fig:1point_v}, \ref{fig:1enseg_v}, \ref{fig:1detecttion}, \ref{fig:1water_v}, \ref{fig:1seg_v}, \ref{fig:1depth_v}, \ref{fig:1normal_v} and \ref{fig:1softedge_v}.

\section{Detailed Editing Instructions}
\label{sec:appendix_prompts}

Table \ref{tab:style_instructions} details the specific text editing instructions used for the qualitative results presented in Figure \ref{fig:style}.

\begin{table}[h!]
\small
\centering
% Table Caption
\caption{Text editing instructions corresponding to the qualitative examples shown in Figure \ref{fig:style}.}
\label{tab:style_instructions}
% 使用 p{12cm} 让长文本自动换行，防止超出边界
\begin{tabular}{c p{12cm}} 
\toprule
\textbf{ID} & \textbf{Text Editing Instruction} \\ 
\midrule
1 & Transformed into 3D Chibi Style, A digital illustration of a young boy and girl, both with big, expressive eyes and wide smiles, standing close together. \\ 
\midrule
2 & Transformed into Flat vector art style. A flat vector illustration of four friends hanging out indoors, sitting on a couch and surrounded by colorful snacks and drinks. \\ 
\midrule
3 & Transformed into low poly style, Photo of two women in a low-quality, pixelated style, with a geometric, faceted appearance. The woman on the left is wearing a black and white outfit with a black bow tie and a white shirt. \\ 
\midrule
4 & Replace the cupcakes with a galaxy theme. \\ 
\midrule
5 & Replace the snowy mountain landscape with a rainbow-colored mountain landscape. \\ 
\bottomrule
\end{tabular}
\end{table}

\end{document}